\documentclass[twoside]{IEEEtran}

\usepackage{graphicx} % Required for inserting images
\usepackage{hyperref}
\usepackage{academicons}
\usepackage{xcolor}
\usepackage{caption}
\usepackage{multirow}
\usepackage{adjustbox}
\usepackage{rotating}
\usepackage{amsmath}
\usepackage{pgfplots}
\pgfplotsset{compat=1.18,width=0cm}
\usepackage{neuralnetwork}
\usepackage{tikz}
\usepackage{CJKutf8}
\usepackage{xcolor}
\usepackage{tcolorbox}
\usepackage[square,sort,comma,numbers]{natbib}
\usepackage{tikz}
\usepackage{CJKutf8}
\usepackage{xcolor}
\usepackage{tcolorbox}
\usepackage{amssymb}
\usepackage{tabularx}
\usepackage{graphicx,subcaption}
\usepackage{lipsum}
\usepackage{authblk}

\begin{document}

%\title{EDformer: Embedded Decomposed Transformer for Explainable Multivariate Time Series Forecasting}

\title{EDformer: Embedded Decomposition Transformer for Interpretable Multivariate Time Series Predictions}

\author{Sanjay Chakraborty, Ibrahim Delibasoglu, Fredrik Heintz }

\affil[]{Department of Computer and Information Science (IDA), REAL, AIICS, Linköping University, Linköping, Sweden, Email: sanjay.chakraborty@liu.se*} 

\maketitle

\begin{abstract}
Time series forecasting is a crucial challenge with significant applications in areas such as weather prediction, stock market analysis, and scientific simulations. This paper introduces an embedded decomposed transformer, 'EDformer', for multivariate time series forecasting tasks. Without altering the fundamental elements, we reuse the Transformer architecture and consider the capable functions of its constituent parts in this work. Edformer first decomposes the input multivariate signal into seasonal and trend components. Next, the prominent multivariate seasonal component is reconstructed across the reverse dimensions, followed by applying the attention mechanism and feed-forward network in the encoder stage. In particular, the feed-forward network is used for each variable frame to learn nonlinear representations, while the attention mechanism uses the time points of individual seasonal series embedded within variate frames to capture multivariate correlations. Therefore, the trend signal is added with projection and performs the final forecasting. The EDformer model obtains state-of-the-art predicting results in terms of accuracy and efficiency on complex real-world time series datasets. This paper also addresses model explainability techniques to provide insights into how the model makes its predictions and why specific features or time steps are important, enhancing the interpretability and trustworthiness of the forecasting results.
\end{abstract}

\textbf{Keywords -} Time-series; Forecasting; Transformer; Multivariate; Explainable AI.

\section{Introduction}
Time-series forecasting plays an important role in a variety of industries, such as sensor network monitoring \citep{li2024deep}, smart grid management \citep{maurya2016time}, economics and finance \citep{zhang2024hybrid}, and disease propagation analysis \citep{wen2022transformers}. By replacing the Long Short-Term Memory (LSTM) and Recurrent Neural Network (RNN) based structure, which has been the main approach for processing time-series data, with several self-attention modules, Transformer shows improved performance, especially in the time series analysis field \cite{atabay2022multivariate}. Transformer is emerging in time series forecasting, driven by the tremendous success in the natural language processing field. Transformer has the ability to extract multi-level representations from sequences and illustrate pairwise connections \cite{zhang2024hybrid}. In time series data analysis, Transformers that combine frequency-domain filtering with temporal and channel-wise transformations offer a powerful approach to capturing complex dependencies and patterns. By leveraging frequency-domain filtering, these models can efficiently separate and analyze the signal's various frequency components, isolating key seasonal or trend-related elements \cite{atabay2022multivariate}. Additionally, temporal and channel-wise transformations allow the model to process both the sequential nature of time series data and the relationships across multiple channels, making it particularly adept at handling multivariate time series and discovering intricate dependencies between variables \cite{ahmed2023transformers}. This dual capability enables the model to address both local and global patterns, improving performance in tasks such as forecasting, anomaly detection, and classification. In addition, the transformer-based model suffers from slow training and inference speed due to the bottleneck incurred by a deep encoder and step-by-step decoder inference. 
\par In this paper, Our main contributions are as follows:
\begin{itemize}
    \item We analyse the Transformer architecture and discover that its native components' efficacy for multivariate time series has not received enough attention.
    \item The EDformer model decomposes the input multivariate time series into separate seasonal and trend components. Each component is then independently embedded into frames using reverse operation, allowing the self-attention module to capture multivariate correlations while the feed-forward network encodes the representations of each series.
    \item EDformer is a lightweight and computationally efficient architecture, that significantly reduces processing time compared to existing state-of-the-art methods while maintaining or even improving forecasting accuracy.
    \item EDformer experimentally reaches cutting-edge performance on real-world benchmarks. A promising direction for future developments in Transformer-based forecasting models is shown by our thorough examination of its embedded modules and architectural decisions.
\end{itemize}
We do comprehensive tests on different time-series forecasting benchmarks to support our motivation and hypothesis. When compared to various forecasting techniques, the performance of our proposed model is at the cutting edge. Extensive ablation investigations and analysis trials support the viability of suggested designs and their consequent benefit over earlier methods.\\
This paper is organized as follows. Section \ref{lit} explains a set of typical times series forecasting works from the literature which sets the notion and motivation of this paper. Section \ref{prob} briefly explains the problem statement of this paper. In Section \ref{Method}, we provide a detailed discussion of our proposed architecture suitable for multivariate time series forecasting. In section \ref{result}, we discuss the result analysis of the proposed methodology and describe a detailed comparison with the state-of-the-art time series forecasting models. A deeper explainability of the proposed EDformer model is described in the section \ref{explain}.  A conclusion of the work's findings is given in section \ref{conclusion}.

\section{Literature Review}
\label{lit}
In long-term and short-term time series forecasting, transformers have gained considerable attention due to their ability to model complex temporal dependencies \cite{zerveas2021transformer,zhang2024multivariate,nie2022time,zeng2023transformers}. Informer \citep{zhou2021informer}, one of the first widely recognized transformers for time-series forecasting, employs a generative-style decoder and ProbSparse self-attention to address the challenge of quadratic time complexity. Subsequently, various transformer-based models have been proposed, including Autoformer \citep{chen2021autoformer}, Pyraformer \citep{liu2022pyraformer}, iTransformer \citep{liu2023itransformer}, Reformer \citep{kitaev2020reformer}, and FEDFormer \citep{zhou2022fedformer}. Pyraformer emphasizes multiresolution attention for efficient signal processing. Autoformer \citep{chen2021autoformer} leverages auto-correlation and decomposition techniques to improve forecasting, while FEDFormer \citep{zhou2022fedformer} integrates frequency-domain analysis to enhance the representation of time series. PatchTST \citep{nie2022time} focuses on the use of patches to improve the model's capacity to capture both local and global dependencies in the data. Crossformer \citep{zhang2023crossformer} introduces the Dimension-Segment-Wise (DSW) embedding method, which encodes input time-series data into a 2D vector array while preserving temporal and dimensional structure. It also employs a Two-Stage Attention (TSA) layer to capture cross-time and cross-dimension dependencies. By combining DSW embedding and TSA, Crossformer builds a Hierarchical Encoder-Decoder (HED) framework to operate across multiple scales for more effective predictions. The Multi-resolution Time Series Transformer (MTST) \citep{zhang2024multi} adopts a multi-branch architecture to model temporal patterns across different resolutions \citep{woo2022etsformer}. MTST distinguishes itself by employing relative positional encoding, which is better suited for capturing periodic components at various scales, compared to the fixed or absolute positional encodings used in many other transformer models.
Time series decomposition, a standard method in time series analysis \citep{cleveland1990stl}, breaks down a time series into distinct components, each reflecting a specific category of patterns that exhibit greater predictability. This technique is particularly effective for examining historical changes over time. In forecasting tasks, decomposition is commonly applied as a pre-processing step for historical data before predicting future trends \citep{asadi2020spatio}. Notable examples include Prophet \citep{sean2018forecasting}, which uses trend-seasonality decomposition, N-BEATS \citep{oreshkin2019n}, which employs basis expansion, and DeepGLO \citep{sen2019think}, which utilizes matrix decomposition. However, such pre-processing approaches are often constrained by the basic decomposition effects of historical series, failing to capture the hierarchical interactions among underlying patterns in long-term forecasts. In this paper, a basic AvgPooling(.) decomposition technique has been used to separate the seasonal and trend patterns from the original input signal.

\section{Problem Statement}
\label{prob}
This paper addresses the challenge of long-term forecasting for multivariate time series, using historical data. We define a multivariate time series at time \( t \) as \( \mathbf{x}_t = [x_{t,1}, x_{t,2}, \dots, x_{t,N}]^\top \in \mathbb{R}^{T \times N} \), where \( x_{t,n} \) represents the value of the \( n \)-th variable at time \( t \), for \( n = 1, 2, \dots, N \). The aim is to develop a model for forecasting the future values of the series over the next \( T \) time steps, based on the most recent \( L \) time steps. The parameters \( L \) and \( H \) are referred to as the \textit{look-back window} and the \textit{prediction horizon}, respectively.
Specifically, for a given initial time \( t_0 \), the model takes as input the sequence \( \mathbf{x}_{t_0-l} \), corresponding to the past \( L \) time steps, and outputs the predicted sequence \( \hat{\mathbf{x}}_{t_0+h} \), representing the forecasted values for the next \( H \) time steps. The predicted value of \( \mathbf{x}_t \) at time \( t \) is denoted by \( \hat{\mathbf{x}}_t \). In brief, the goal of multivariate time series forecasting is to predict future values \( \mathbf{X}_{t+h} \in \mathbb{R}^{T \times N} \) given past observations \( \mathbf{X}_{t-l} \):
\begin{equation}
\hat{\mathbf{X}}_{t+h} = f(\mathbf{X}_{t-l})
\end{equation}
The forecasting performance of the model is assessed by computing the mean squared error (MSE) and the mean absolute error (MAE) between the prediction and the ground truth on the test set:
\begin{equation}
\text{MSE} = \frac{1}{N} \sum_{i=1}^{n} \left\| \mathbf{X}_{t+h}^{(i)} - \hat{\mathbf{X}}_{t+h}^{(i)} \right\|^2
\end{equation}
\begin{equation}
\text{MAE} = \frac{1}{N} \sum_{i=1}^{n} \left\| \mathbf{X}_{t+h}^{(i)} - \hat{\mathbf{X}}_{t+h}^{(i)} \right\|
\end{equation}

\section{Methodology}
\label{Method}
\begin{figure*}[hbt!]
\centering
\includegraphics[scale=0.45]{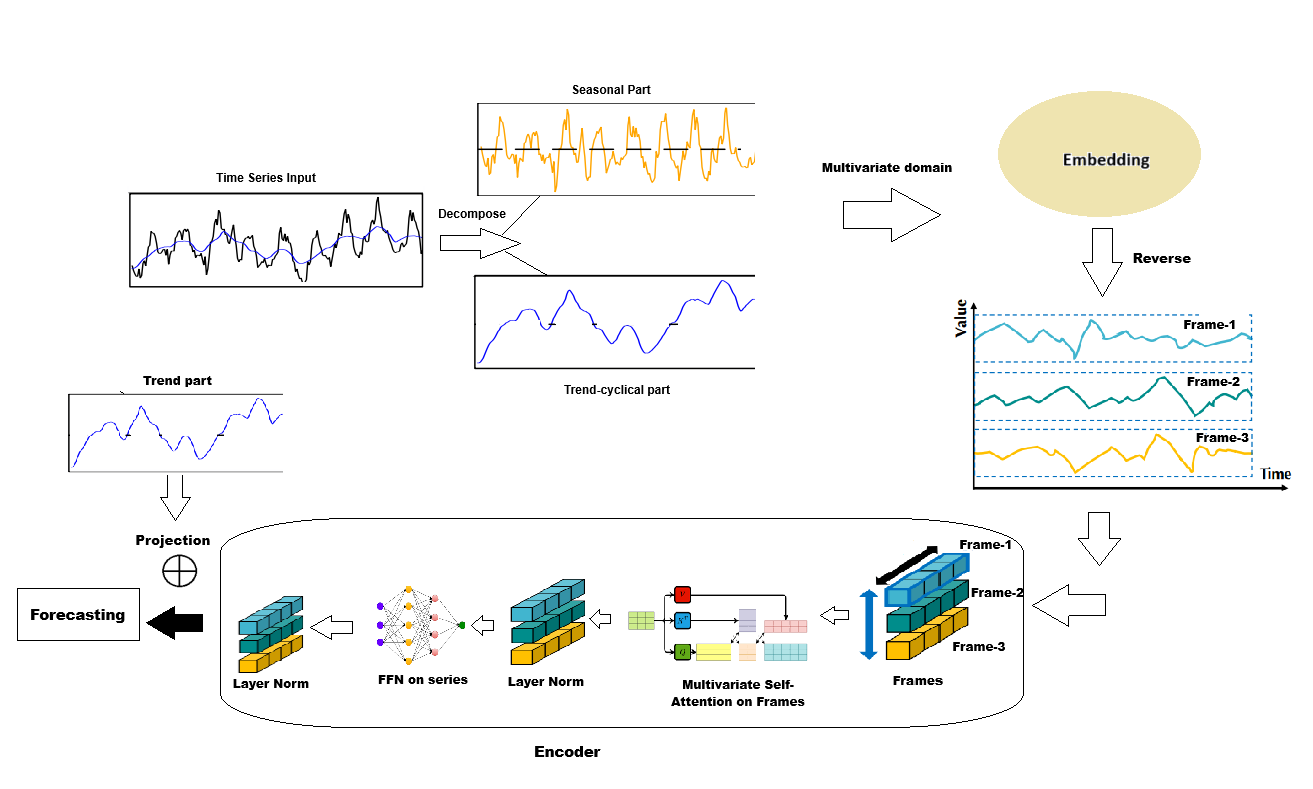}
\caption{Overall architecture of EDformer}
\label{fig-EDformer_architecture}
\end{figure*}

In this work, we have discussed the proposed methodology with a deep decomposition architecture, which includes a series decomposition block, reverse operation, embedding operation, and corresponding encoder. Our proposed EDformer architecture uses transformer's encoder-only architecture. The general overview of the proposed EDformer architecture is represented in Figure \ref{fig-EDformer_architecture}. In time-token-based transformers, the embedding of points from the same time step, which fundamentally represent distinct physical meanings captured by inconsistent measurements, results in the loss of multivariate correlations within a single token. Such tokens, formed from isolated time steps, face challenges in extracting useful information due to their overly local receptive fields and the misalignment of events across simultaneous time points. Furthermore, while the order of sequences plays a critical role in influencing time series variations, the use of permutation-invariant attention mechanisms on the temporal dimension is unsuitable, as it disregards the sequential nature of the data \cite{zeng2023transformers}.

\subsection{Decomposition Block}
Decomposition has been a widely used method in time series analysis for many years \citep{ahmed2023transformers}. The decomposition block separates a time series into its seasonality and trend-cycle components, enhancing the model's ability to accurately capture these aspects. In time series analysis, decomposition involves breaking down a time series into three systematic components: trend-cycle, seasonal variation, and random fluctuations. The trend component reflects the long-term direction of the series, which may be increasing, decreasing, or stable over time. The seasonal component captures recurring patterns within the series, while the random (or "irregular") component accounts for noise that cannot be explained by the trend or seasonal factors. Decomposition can be performed in two main ways: additive and multiplicative. By breaking down a time series into these components, we gain a deeper understanding of the underlying patterns, allowing for more effective modelling. As can be seen, the encoder and decoder use a decomposition block to aggregate the trend-cyclical part and extract the seasonal part from the series progressively. Example, let us take an input series $X\in R^{L \times D}$ with length L, and the decomposition layer returns $X_{T}$ and $X_{S}$ defined as,
\begin{equation}
\begin{split}
X_{T}=AvgPooling(Padding(X)) \\
X_{S}=X-X_{T}
\end{split}
\end{equation}
$X_{S}$ and $X_{T}$ denote the seasonal and the extracted trend-cyclical parts, respectively. We use the AvgPool(.) for the moving average with the padding procedure to maintain the series length. Applying a reverse time series transformer (EDformer) on decomposed signals, specifically on the seasonal and trend components separately, enhances model interpretability and precision. By isolating these components, the model can better capture distinct temporal patterns and dependencies within each, leading to improved accuracy in detecting seasonality and trends. This approach also reduces noise interference, allowing the transformer to focus on key features in each decomposed signal. Consequently, the model achieves a more nuanced representation of the underlying data dynamics than when applied to the whole signal at once.
\subsection{Reversible Model Inputs and Embedding}
\begin{figure}[hbt!]
\centering
\includegraphics[scale=0.3]{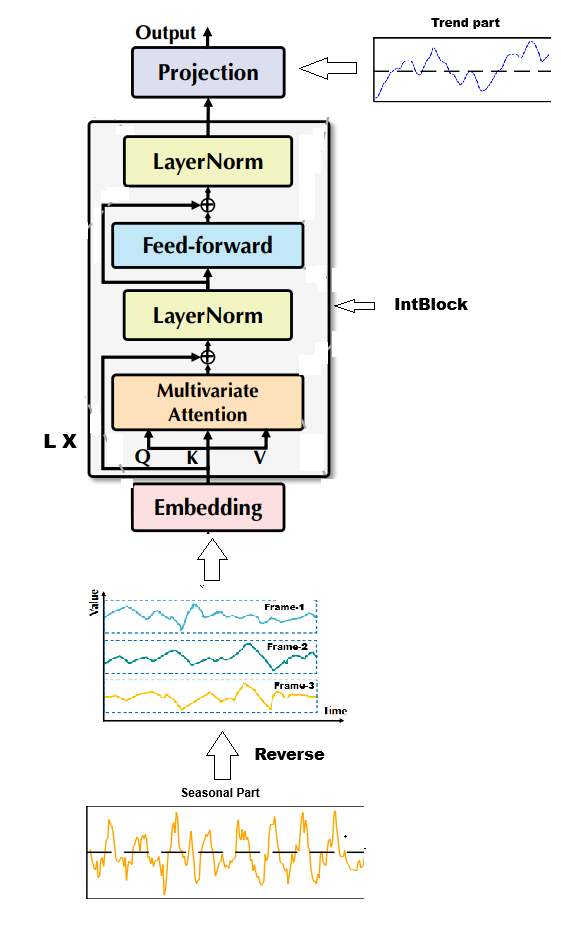}
\caption{Internal architecture of EDformer blocks}
\label{Fig2}
\end{figure}

Instead of taking multiple temporal tokens, our approach takes one whole signal (Seasonal and Trends components) as a single frame. The concept of processing the entire signal within a single frame is inspired by the iTransformer model \citep{liu2023itransformer}. Across multivariate series, our proposed encoder-only EDformer architecture encourages adaptive correlation and representation learning. Every time series is first tokenized into a frame  capture the distinct characteristics of every variable from both the components (seasonal+trend). After modeling mutual interactions with self-attention, feed-forward networks process each series (seasonal+trends) separately to provide its representation. This framework is optimized for capturing intricate temporal dependencies in time series. The procedure of forecasting future series of each distinct frame $Y^t:,n$ based on the lookback series $X^S:,n$ and $X^T:,n$ in EDformer is straightforwardly expressed as follows in light of the aforementioned considerations,
\begin{equation}
\begin{split}
h^0_n=Embedding(Reverse(X^S:,n))\\ + Embedding(Reverse(X^T:,n)),\\
H^{(l+1)}=IntBlock(H^l), l= 0, ....., L-1, \\
Y^t:,n = Projection(h^L_n),
\end{split}
\end{equation}
where the superscript indicates the layer index and $H = {h_1,..., h_N } \in R^{N×D}$ contains N embedded tokens of size D. Multi-layer perceptrons (MLPs) are used to implement both embedding and projection. The shared feed-forward network in each IntBlock() processes the acquired frames individually while interacting with one another through self-attention. In particular, the position embedding in the vanilla Transformer is no longer required in this case because the neuron permutation of the feed-forward network automatically stores the order of sequence. The internal architecture of EDformer is represented
in Figure \ref{Fig2}.

\subsection{Model Encoding}
In the encoder block, we arrange a stack of L blocks made up of the feed-forward network, self-attention, and layer normalization modules. 

\subsubsection{Multivariate Self-attention}
The proposed model views the entire series of a single variable as an independent process, whereas the attention mechanism is typically used to facilitate the modelling of temporal connections in forecasting. The self-attention mechanism allows the model to weigh the importance of different tokens in a sequence relative to each other, facilitating the capture of long-range dependencies and contextual relationships. It adopts linear projections to obtain queries(Q), Keys(K) and values(V),
\begin{equation}
Q=SW_Q, K=SW_k,V=SW_v
\end{equation}
Where, $W_Q, W_k, W_v$ are learned weight matrices. Several keys, queries, and values are fed into multi-scaled dot-product attention blocks by multi-head attention, which then concatenates the attention to produce the desired result. The attention scores can be computed as,
\begin{equation}
scores=\frac{QK^T}{\sqrt{d_k}}
\end{equation}
where n represents an input sequence of length and $d_k$ represents the projected dimension. The computational complexity of computing the attention score is $O(n^2d)$. Therefore, we execute Softmax normalization ($O(n^2d)$) and weighted sum approaches.
\begin{equation}
Attention(Q,K,V)=Softmax(scores)
\end{equation}
\begin{equation}
Output=Attention(Q,K,V)*V
\end{equation}
Transformer uses multivariate self-attention (MVA) with multiple sets of learned projections (H different time series) instead of a single attention function.
\begin{equation}
MVA(Q,K,V)=Concat(h_1,....,h_H)W^O
\end{equation}
Where, $h_i=Attention(QW_i^Q,KW_i^K,VW_i^V)$.

\subsubsection{Layer Normalization}
The "Layer Normalization (LayerNorm)" block comes next \citep{ba2016layer}. Initially, layer normalisation was suggested as a way to improve deep networks' training stability and convergence. The module in most Transformer-based forecasters gradually fuses the variables with one another by normalising the multivariate representation of the same timestamp. Our inverted version applies normalisation to the individual variate's series representation as Equation \ref{eq11}, which has been researched and shown to be useful in solving non-stationary situations. On the other hand, an oversmooth time series will result from the normalisation of various tokens of time steps in the prior design.
\begin{equation}
\label{eq11}
LayerNorm(H) = \left[  \frac{ h_n - Mean(h_n) } { \sqrt{Var(h_n)} } | n = 1, ....., N \right]
\end{equation}

\subsubsection{Feed-forward network}
The feed-forward network (FFN) is used by the transformer as the fundamental building block for token representation encoding, and it is applied consistently to every token. As previously indicated, several variations of the same timestamp that make up the token in the vanilla Transformer may be malpositioned and too localised to provide sufficient information for predictions. FFN is used on each variate frame's series representation in the reversed form. They are focused on storing the observed decomposed time series and decoding the representations for subsequent series utilising dense non-linear connections by stacking inverted blocks \citep{das2023long}. As a more effective predictive representation learner than self-attention applied on time points, a logical explanation has been provided in which the neurones of MLP are trained to depict the intrinsic properties of any time series, such as the amplitude, periodicity, and even frequency spectrums. This is sent to a "feed-forward (FFN)" block, the output of which has a "LayerNorm" block. In the encoder block, the entire multi-head attention and feed-forward blocks are repeated n number of times.
\begin{equation}
FFN(H')=ReLU(H'W^1+b^1)W^2+b^2
\end{equation}
Here, $H'$ is an output of the previous layer, $W_1,W_2,b^2$ are trainable parameters. In a deeper module, a residual connection module followed by a layer normalization module is inserted around
each module. 
\begin{equation}
H'=LayerNorm(MVSelfAtten(X)+X)
\end{equation}
\begin{equation}
H=LayerNorm(FFN(H')+H')
\end{equation}
Where MVSelfAtten(.) defines the multivariate self-attention module and LayerNorm(.) denotes the layer normalization task. 

\subsection{EDformer Components}
The entire architecture of the proposed EDformer model is discussed below,\\\\
\textbf{1. Decomposition:}\\
The decomposition layer decomposes the input time series data into trend and seasonality components. The $moving_{avg}()$ layer is used to capture the trend, and the residual (difference between the input and trend) represents the seasonality.\\\\
\textbf{2. Model:}\\
The model sets up the structure, including:\\
\textit{2.1 Embedding Layer:} Converts the time series data into a form suitable for the transformer encoder.\\
\textit{2.2 Transformer Encoder:} Includes multiple layers of self-attention, allowing the model to capture dependencies within the time series.\\\\
\textbf{3. Forecasting Process:}\\
This method performs the forecasting process by:\\
- Normalizing seasonality to improve stability.\\
- Encoding the normalized seasonality with embeddings and transformer layers.\\
- Projecting and de-normalizing the output to make it compatible with the prediction.\\
- Adding the trend component back for the final forecast.\\\\
\textbf{4. Forward Process:}
Handles forward propagation, using the forecast method for forecasting tasks.

\section{Result Analysis}
\label{result}
\subsection{Datasets}

The datasets used in this study, detailed in Table \ref{table-dataset}, encompass both long-term and short-term time series forecasting scenarios. The 'Electricity Transformer Temperature (ETT)' dataset contains 7 factors of electricity transformer measurements spanning with four subsets: ETTh1 and ETTh2 recorded hourly, and ETTm1 and ETTm2 recorded every 15 minutes. The Exchange dataset comprises daily exchange rates from 8 countries between 1990 and 2016. Additional datasets include 'Weather' (21 meteorological factors collected every 10 minutes), 'Electricity Consumption Load (ECL)' with hourly data from 321 clients, and Traffic data consisting of hourly road occupancy rates from 862 sensors in the San Francisco Bay area. The Solar-Energy dataset records 10-minute samples of solar power production from 137 PV plants. PEMS traffic data is widely associated with the California 'Performance Measurement System (PeMS)', a comprehensive database used for freeway performance monitoring in California. For short-term forecasting, we utilize four subsets of the PEMS traffic network data (PEMS03, PEMS04, PEMS07, and PEMS08), which contain traffic flow information recorded at 5-minute intervals \citep{pems_dataset}. These diverse datasets, with varying temporal granularities and dimensionalities, enable comprehensive evaluation of forecasting models across different domains and time scales. The M4 experimental dataset consists of 100,000 time series covering diverse domains, as detailed in Table \ref{M4dataset}. This M4 dataset allows us to evaluate the proposed EDformer model's ability to handle short-term prediction tasks with high variability. All datasets are publicly available and divided into training, validation, and test sets in the benchmark Time-Series Library (TSLib).

\begin{table*}[hbt!]
\centering
\scriptsize
\captionsetup{justification=centering}
\caption{Detailed dataset descriptions for long-term and short-term forecasting datasets}
\begin{tabular}{|c|c|c|c|c|c|}
\hline
Forecasting & Dataset      & Dim & Size                & Frequency & Information    \\ \hline
Long-term   & ETTh1, ETTh2 & 7   & (8545,2881,2881)    & Hourly    & Electricity    \\
            & ETTm1, ETTm2 & 7   & (34465,11521,11521) & 15 min    & Electricity    \\
            & Weather      & 21  & (36792,5271,10540)  & 10 min    & Weather        \\
            & Electricity  & 321 & (18317,2633,5261)   & Hourly    & Electricity    \\
            & Traffic      & 862 & (12185,1757,3509)   & Hourly    & Transportation \\
            & Exchange     & 8   & (5120,665,1422)     & Daily     & Economy        \\ \hline
Short-term  & PEMS03       & 358 & (15617,5135,5135)   & 5 min     & Transportation \\
            & PEMS04       & 307 & (10172,3375,3375)   & 5 min     & Transportation \\
            & PEMS07       & 883 & (16911,5622,5622)   & 5 min     & Transportation \\
            & PEMS08       & 170 & (10690,3548,3548)   & 5 min     & Transportation \\ \hline
\end{tabular}
\label{table-dataset}
\end{table*}

\begin{table*}[hbt!]
\centering
\scriptsize
\captionsetup{justification=centering}
\caption{Number of M4 series per data frequency and domain}
\begin{tabular}{|c|c|c|c|c|c|c|c|}
\hline
Time interval between successive observations & Micro           & Industry        & Macro           & Finance         & Demographic    & Other          & Total            \\ \hline
Yearly                                        & 6,538           & 3,716           & 3,903           & 6,519           & 1,088          & 1,236          & \textbf{23,000}  \\ \hline
Quarterly                                     & 6,020           & 4,637           & 5,315           & 5,305           & 1,858          & 865            & \textbf{24,000}  \\ \hline
Monthly                                       & 10,975          & 10,017          & 10,016          & 10,987          & 5,728          & 277            & \textbf{48,000}  \\ \hline
Weekly                                        & 112             & 6               & 41              & 164             & 24             & 12             & \textbf{359}     \\ \hline
Daily                                         & 1,476           & 422             & 127             & 1,559           & 10             & 633            & \textbf{4,227}   \\ \hline
Hourly                                        & 0               & 0               & 0               & 0               & 0              & 414            & \textbf{414}     \\ \hline
Total                                         & \textbf{25,121} & \textbf{18,798} & \textbf{19,402} & \textbf{24,534} & \textbf{8,708} & \textbf{3,437} & \textbf{100,000} \\ \hline
\end{tabular}
\label{M4dataset}
\end{table*}

\subsection{Experimental Analysis}
\begin{figure}[hbt!]
  \begin{subfigure}{0.9\columnwidth}
  \includegraphics[width=\textwidth]{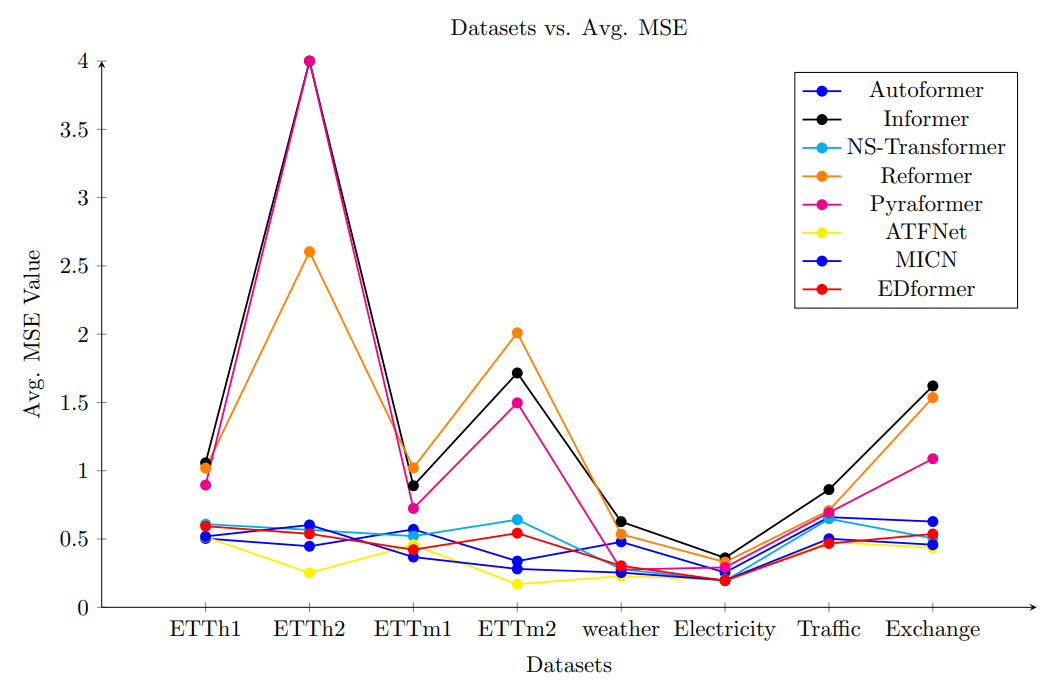}
  %\caption{Image A}
  \end{subfigure}
  \hfill
  \begin{subfigure}{0.9\columnwidth} 
  \includegraphics[width=\textwidth]{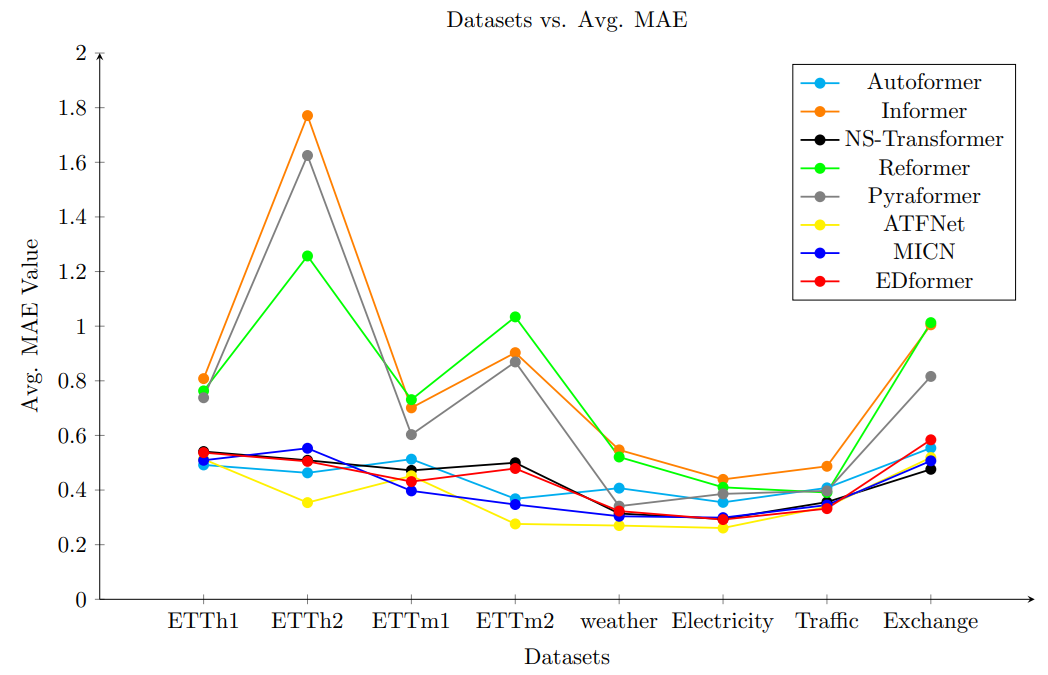} 
  %\caption{Image B} 
  \end{subfigure}
  \caption{Comparison of models efficiency with datasets vs. avg. MSE vs. avg. MAE}
  \label{fig-comparison}
  \end{figure}

\begin{figure}[hbt!]
  \begin{subfigure}{0.49\columnwidth}
  \includegraphics[width=\textwidth]{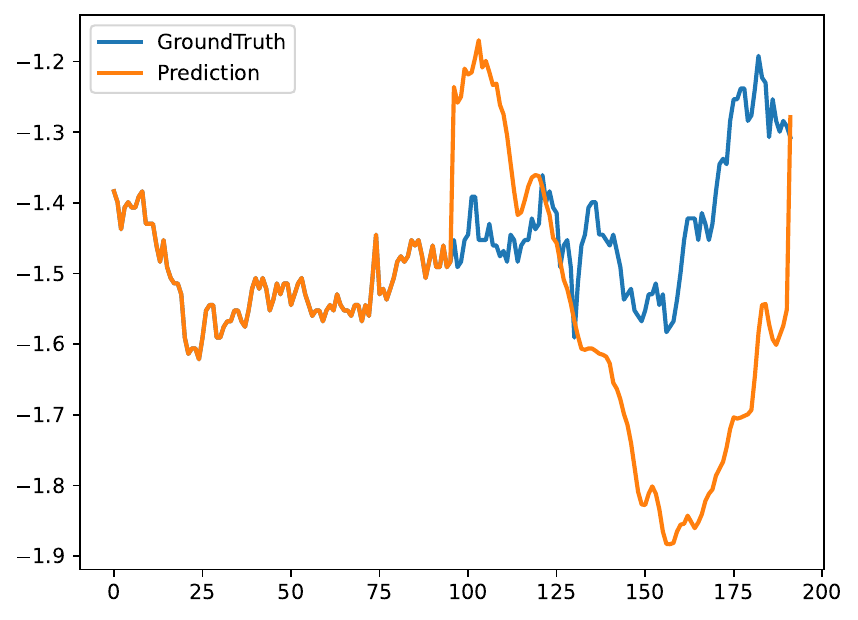}
  \caption{Autoformer} 
  \end{subfigure} 
  \hfill
  \begin{subfigure}{0.49\columnwidth} 
  \includegraphics[width=\textwidth]{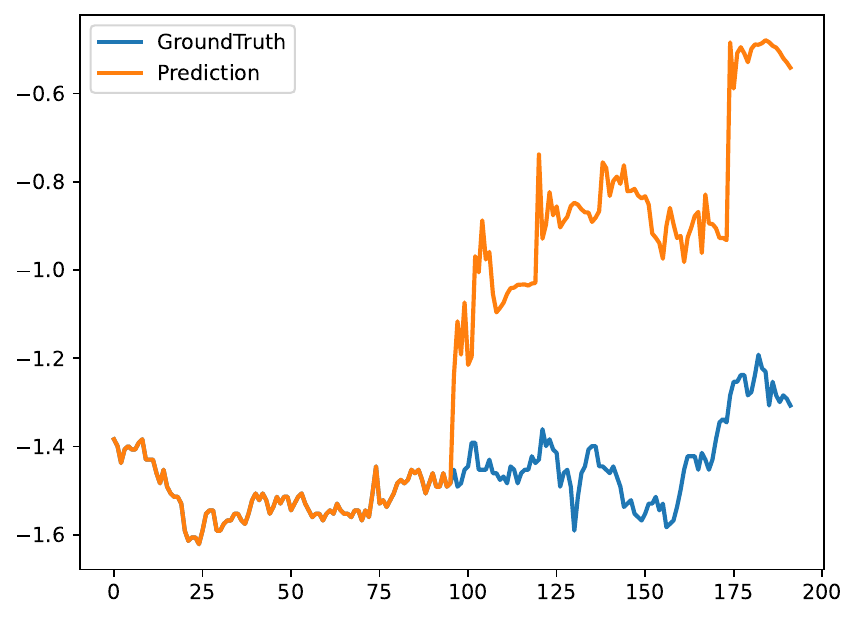} 
  \caption{Informer} 
  \end{subfigure}
  \begin{subfigure}{0.49\columnwidth}
  \includegraphics[width=\textwidth]{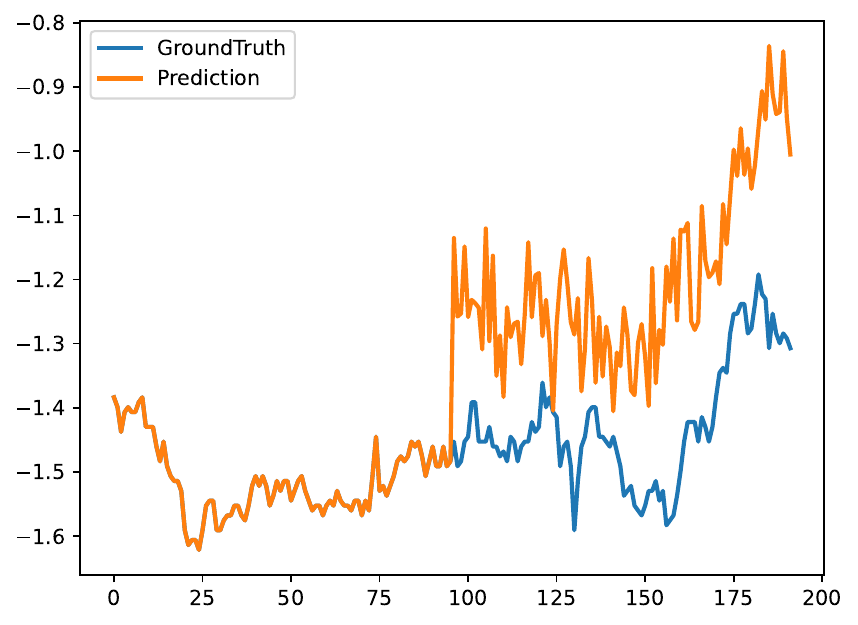}
  \caption{Pyraformer}
  \end{subfigure} 
  \hfill 
  \begin{subfigure}{0.49\columnwidth} 
  \includegraphics[width=\textwidth]{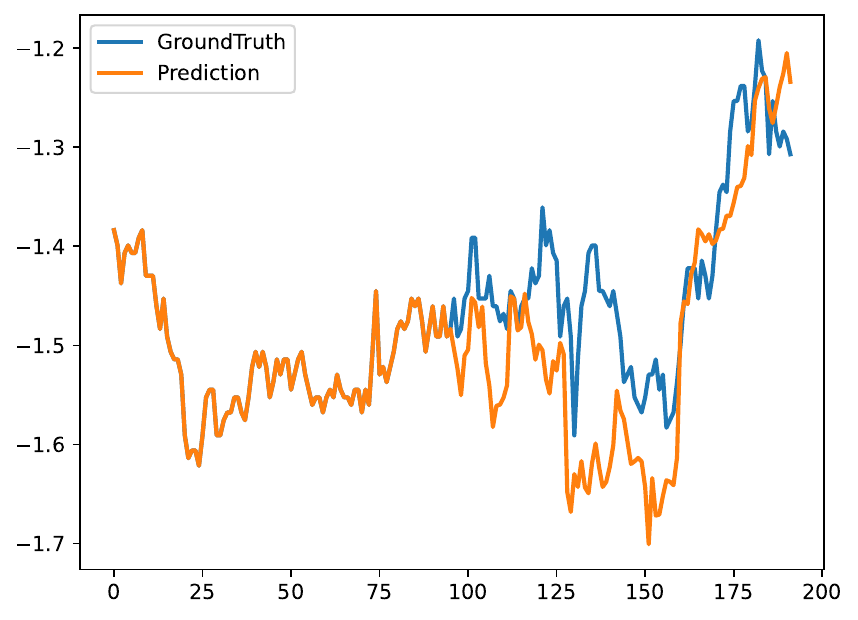} 
  \caption{MICN} 
  \end{subfigure}  
  \hfill
  \begin{subfigure}{0.49\columnwidth} 
  \includegraphics[width=\textwidth]{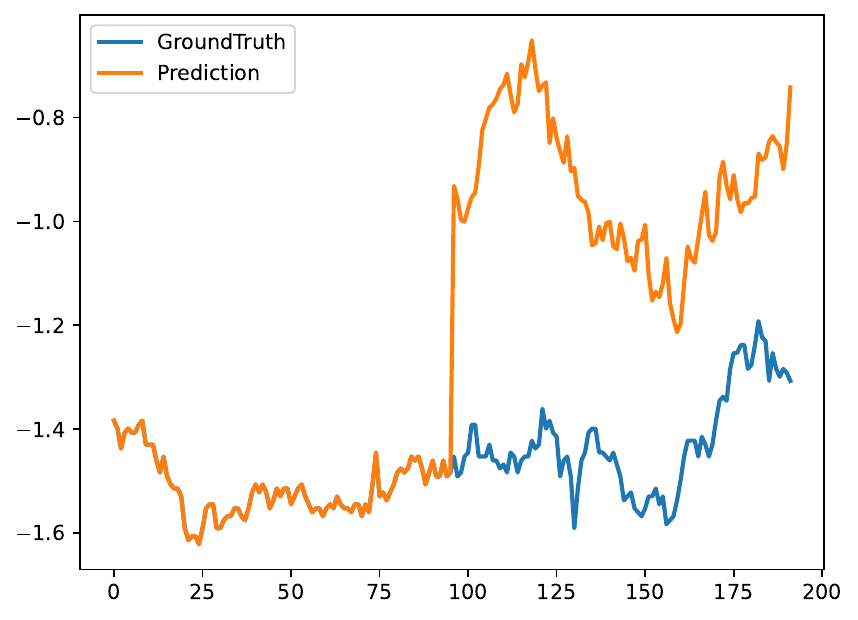} 
  \caption{Reformer} 
  \end{subfigure}  
  \hfill
  \begin{subfigure}{0.49\columnwidth} 
  \includegraphics[width=\textwidth]{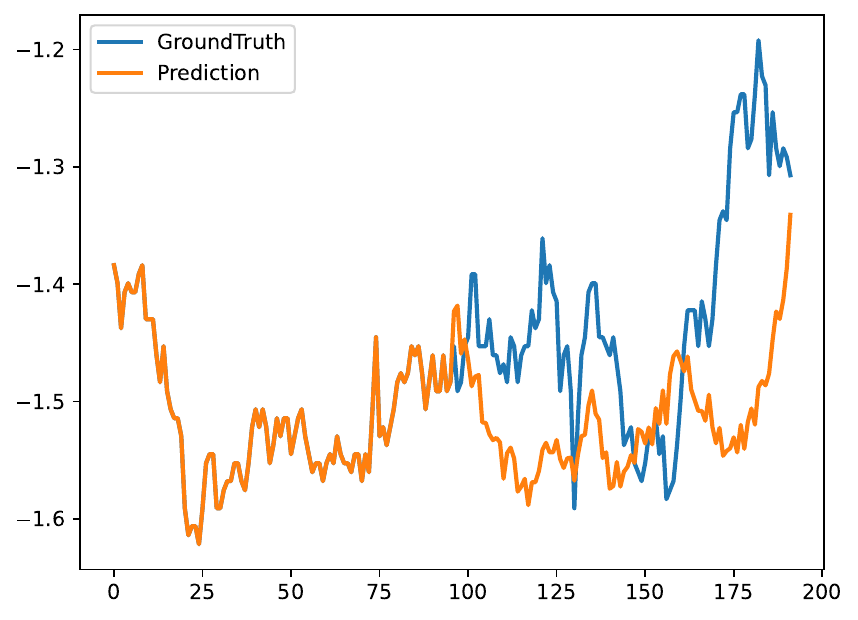} 
  \caption{EDformer} 
  \end{subfigure} 
   \hfill
   
  \caption{Visualization of predictions (length:96) on ETTm1 dataset}
  \label{predtest1}
\end{figure}

\begin{figure}[hbt!]
  \begin{subfigure}{0.49\columnwidth}
  \includegraphics[width=\textwidth]{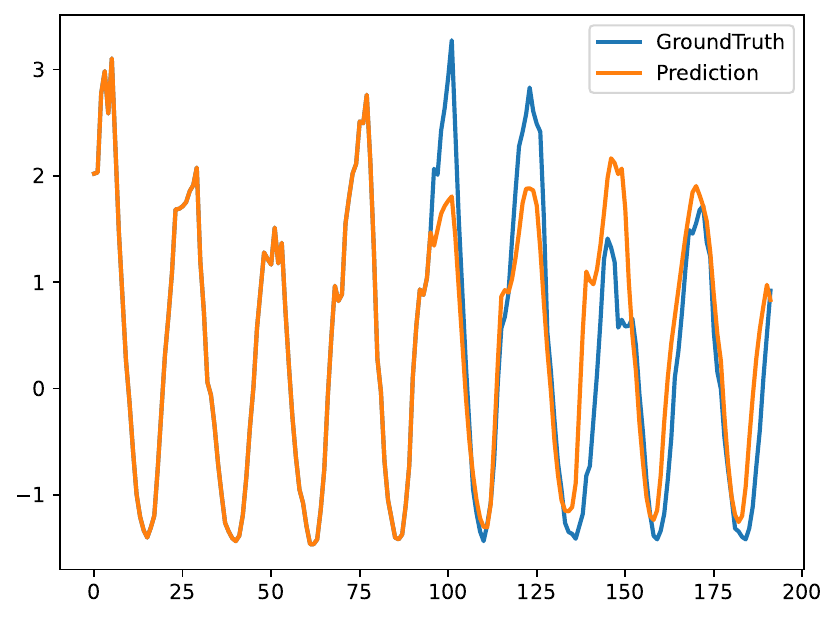}
  \caption{Autoformer} 
  \end{subfigure} 
  \hfill
  \begin{subfigure}{0.49\columnwidth} 
  \includegraphics[width=\textwidth]{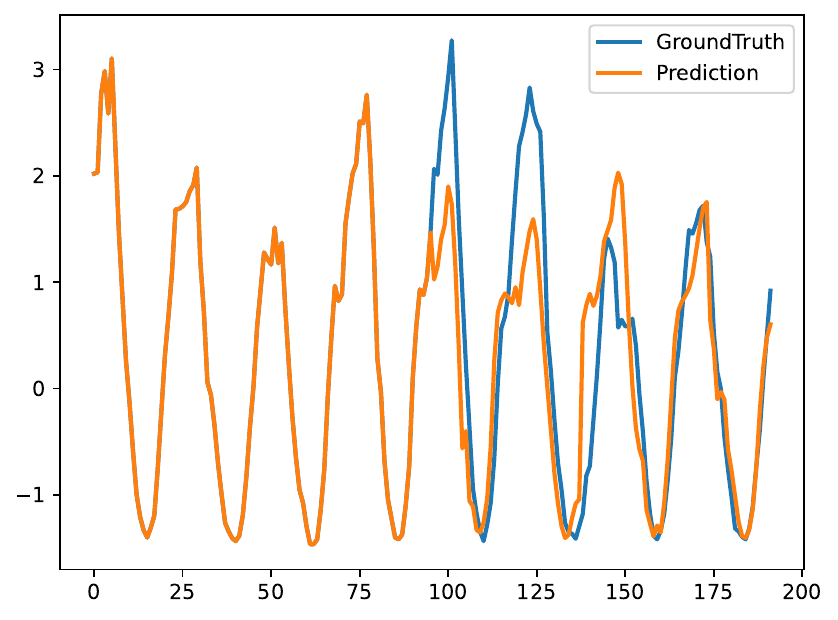} 
  \caption{Informer} 
  \end{subfigure}
  \begin{subfigure}{0.49\columnwidth}
  \includegraphics[width=\textwidth]{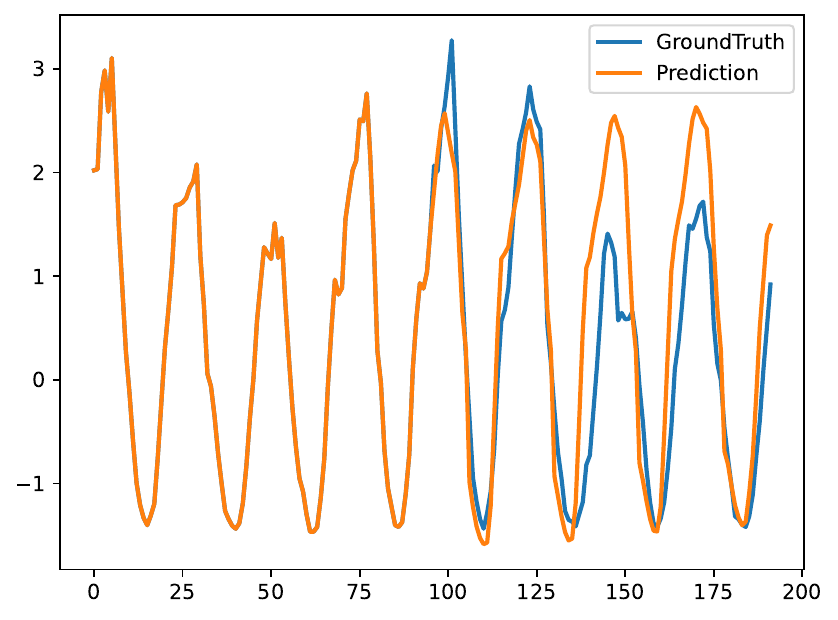}
  \caption{NSTransformer}
  \end{subfigure}
  \hfill
  \begin{subfigure}{0.49\columnwidth} 
  \includegraphics[width=\textwidth]{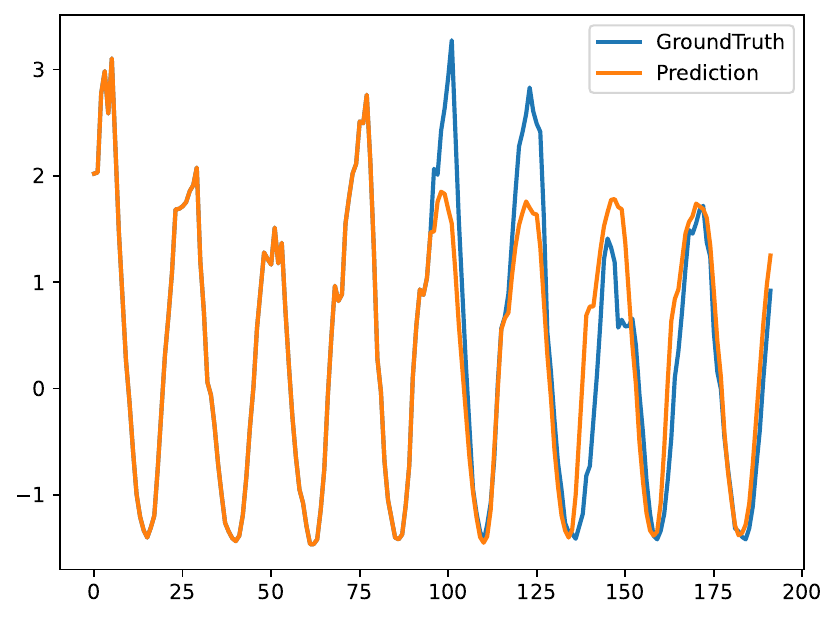} 
  \caption{MICN} 
  \end{subfigure}  
  \hfill
  \begin{subfigure}{0.49\columnwidth} 
  \includegraphics[width=\textwidth]{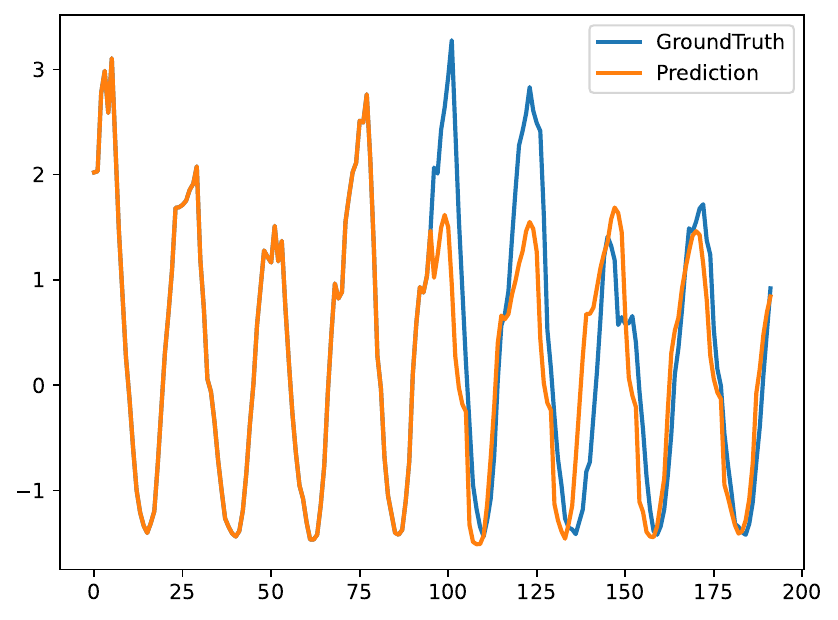} 
  \caption{Reformer} 
  \end{subfigure} 
  \hfill
  \begin{subfigure}{0.49\columnwidth} 
  \includegraphics[width=\textwidth]{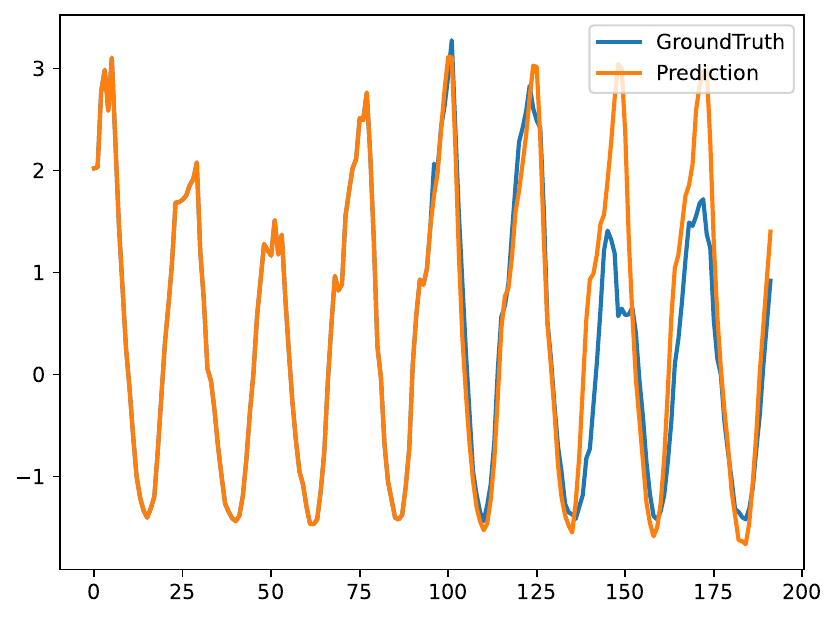} 
  \caption{EDformer} 
  \end{subfigure}  
  \hfill 
  
  \caption{Visualization of predictions (length:96) on Traffic dataset}
  \label{predtest2}
\end{figure}

\begin{figure}[hbt!]
  \begin{subfigure}{0.49\columnwidth}
  \includegraphics[width=\textwidth]{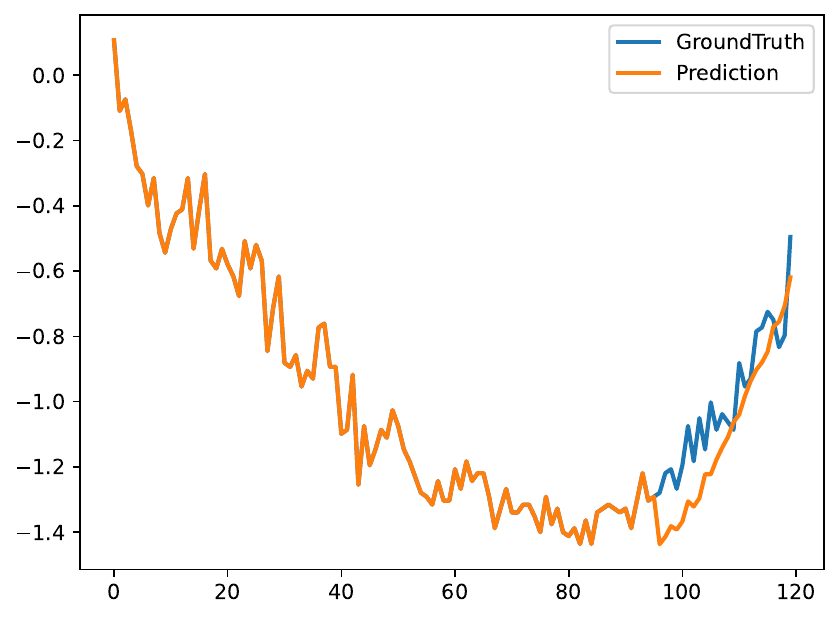}
  \caption{EDformer} 
  \end{subfigure} 
  \hfill
  \begin{subfigure}{0.49\columnwidth} 
  \includegraphics[width=\textwidth]{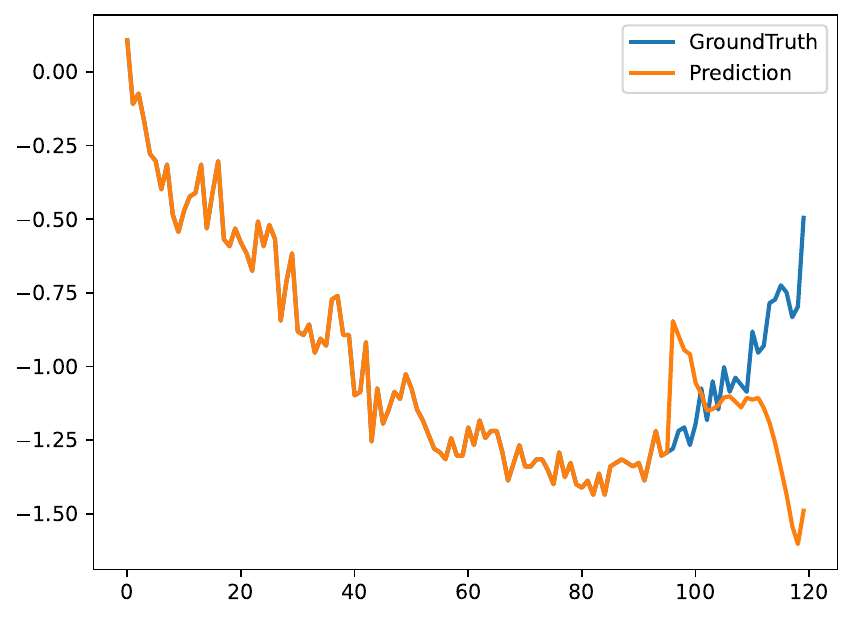} 
  \caption{Autoformer} 
  \end{subfigure}
  \begin{subfigure}{0.49\columnwidth}
  \includegraphics[width=\textwidth]{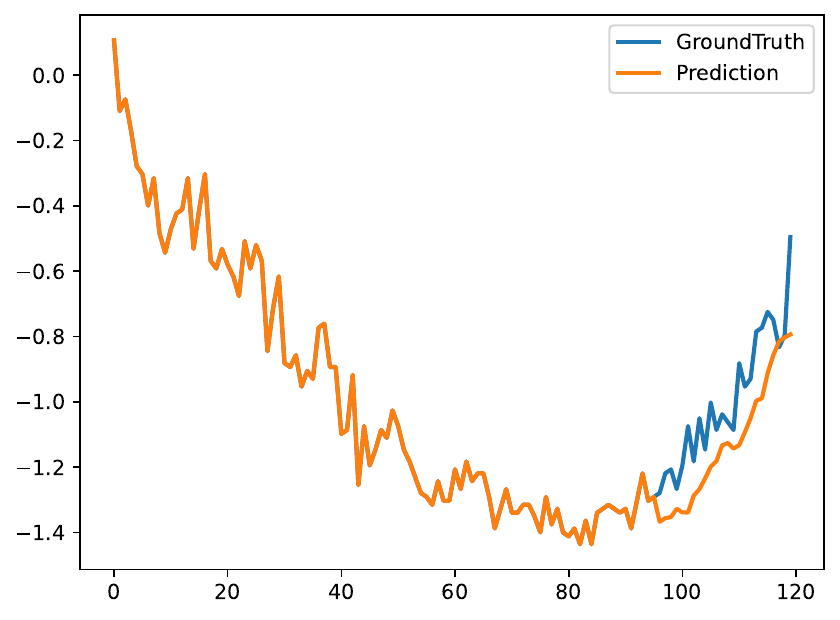}
  \caption{Informer}
  \end{subfigure}
  \hfill
  \begin{subfigure}{0.49\columnwidth} 
  \includegraphics[width=\textwidth]{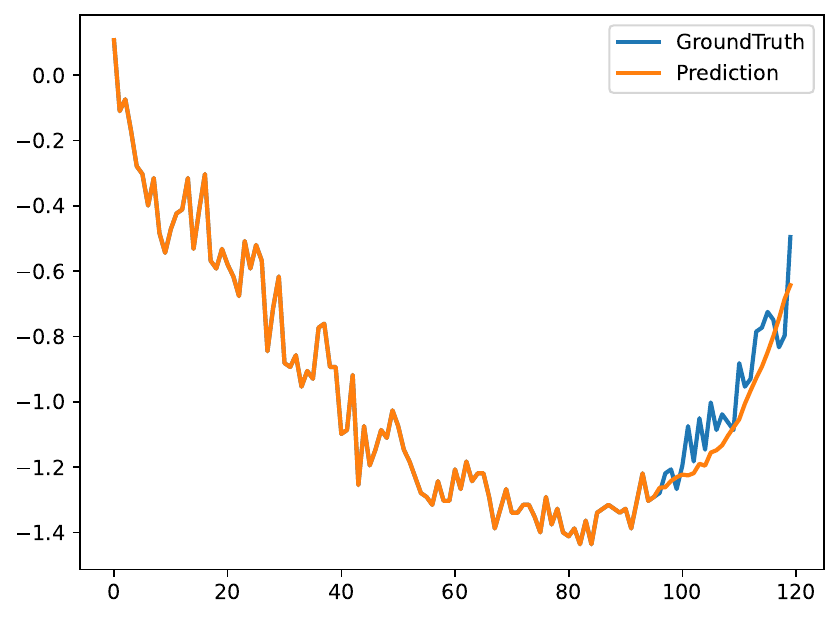} 
  \caption{Pyraformer} 
  \end{subfigure}  
  \hfill 
  
  \caption{Visualization of prediction results on PEMS03 dataset for prediction length 24}
  \label{fig-PEMS03-compare}
\end{figure}

In this section, we perform in-depth experiments to assess how well our proposed EDformer model forecasts in conjunction with state-of-the-art time-series forecasting architectures. An ablation work is also applied to measure the effect of the proposed modules. All the experiments are implemented in PyTorch, CUDA version 12.2 and conducted on a single NVIDIA-GeForce RTX 3090 with 24GB GPU. We have replicated all of the compared baseline models and implemented them using the benchmark Time-Series Library (TSLib) \citep{sa_timeseries} repository, which is based on the configurations provided by the official code or actual article for each model. We train the proposed model on all datasets using a batch size of $32$, a learning rate of $0.0001$ except PEMS, and the ADAM optimizer with L2 loss. The model operates with $K=4$ scales. \\
The comprehensive forecasting results are listed in Tables \ref{tab1}, \ref{tab2}, and \ref{avgtab}. The inclusion of an embedded reverse decomposition process further optimizes the model's performance. A lower MSE/MAE reflects more accurate predictions, and our proposed lightweight EDformer consistently achieves the best performance across most datasets for multivariate long-term forecasting analysis. It outperforms state-of-the-art models such as Autoformer \citep{chen2021autoformer}, Informer \citep{zhou2021informer}, Reformer \citep{kitaev2020reformer}, Pyraformer, Nonstationary-transformer (NS-Trans \citep{liu2022non}), ATFNet \citep{ye2024atfnet}, and MICN \citep{wang2023micn}, particularly excelling in high-dimensional time series forecasting. Figure \ref{predtest1} and \ref{predtest2} show sample long-term predictions for some popular architectures: EDformer, Autoformer, Informer, MICN, Nonstationary transformer and Pyraformer on ETTm1 and traffic datasets. In addition to the results presented in Table \ref{avgtab}, further comparison of forecasting methods, including models listed in the table, is provided in Figure \ref{fig-comparison}. Additionally, a comparison of these widely used models based on computational cost, speed, and average execution time is provided in Tables \ref{tabspeed} and \ref{table-executionpems}. For short-term forecasting performance, the proposed model has competitive performance with Autoformer, Informer, MICN, Non-stationary transformer (NS-Trans) and Pyraformer as shown in Table \ref{avgtabshort}. The additional experiment using the M4 dataset is presented in Table \ref{table-resultsM4}. Table \ref{table-resultsM4} shows that our proposed EDformer model performs well and provides state-of-the-art accuracy compared to other models. Table \ref{tabspeed} presents a comparison of multivariate short-term forecasting models in terms of execution time (in seconds) across four PEMS datasets. Figure \ref{fig-PEMS03-compare} shows sample short-term predictions for some popular architectures: EDformer, Autoformer, Informer, and Pyraformer on the PEMS03 dataset. From the perspective of the PEMS dataset, our EDformer model competes strongly with state-of-the-art models in terms of accuracy. Additionally, EDformer stands out for its lightweight design, achieving comparable results in significantly less time compared to other approaches. However, an extensive analysis of standard deviations on the long-term forecasting values (MSE, MAE) of each dataset and horizon is described in Table \ref{stdlong}. A moderate-to-low standard deviation is desirable. It indicates that the time series has predictable patterns, and the forecasting model can achieve low MSE and MAE.

\begin{table*}[hbt!]
\centering
\scriptsize
\captionsetup{justification=centering}
\caption{Measure of error coefficients on multivariate long-term forecasting results with different prediction lengths (96, 192, 336, 720)}
\begin{tabular}{|p{1.7cm} |c |c c |c c |c c |c c |}
\hline
Models & &\multicolumn{2}{c}{Autoformer} &\multicolumn{2}{c}{Informer} &\multicolumn{2}{c}{NS-Trans} &\multicolumn{2}{c}{Reformer} \\
\hline
Database &Metric  &MSE &MAE &MSE &MAE  &MSE &MAE  &MSE &MAE  \\
\hline
ETTh1 &96       & 0.505& 0.482      &0.649 &0.554 &0.574 &0.513 &0.857 &0.689   \\
    &192        & 0.477& 0.471      & 1.012& 0.786   &0.941 &0.728   &0.843&0.714    \\
    &336        & 0.525&0.507          & 1.029& 0.782       &0.633 &0.558  &1.026 &0.772   \\
    &720        & 0.512&0.509          & 1.240& 0.892    &0.676 &0.584  &1.255 &0.866   \\
 \hline 
  % \textbf{Avg} &   & 0.504& 0.492     &1.058&0.808  &0.609&0.541   &1.019&0.763   \\
 %\hline
ETTh2 &96         & 0.375          & 0.409       &2.835&1.329   &0.484&0.461      &2.202&1.143    \\
        &192        & 0.463          & 0.461       &6.161&2.079  &0.549&0.500   &2.699&1.283   \\
        &336         & 0.472          & 0.494        &5.372&1.951   &0.593&0.522    &2.767&1.266    \\
        &720          & 0.481          & 0.489         &4.292&1.728   &0.644&0.556    &2.749&1.338   \\
 \hline 
  % \textbf{Avg} &   & 0.447          & 0.463      &4.665&1.771  &0.567&0.509   &2.604&1.257   \\
 %\hline
ETTm1  &96     & 0.539          & 0.494         &0.627&0.560  &0.423&0.416   &0.881&0.663 \\
        &192    & 0.566          & 0.511        &0.727&0.618  &0.476&0.450  &0.876&0.682  \\
        &336     & 0.621          & 0.534        &1.248&0.891  &0.579&0.491  &1.063&0.750  \\
        &720     & 0.561          & 0.515      &0.959&0.737  &0.609&0.533  &1.264&0.830 \\
 \hline 
   %\textbf{Avg} &    & 0.571          & 0.513    &0.890&0.701  &0.521&0.472   &1.021&0.731   \\
 %\hline
ETTm2  &96    & 0.295          & 0.337     &0.372&0.459  &0.256&0.318  &0.718&0.630  \\
        &192    & 0.292          & 0.349     &0.819&0.711  &0.570&0.463  &1.808&1.003  \\
        &336   & 0.329          & 0.365      &1.382&0.902  &0.614&0.503  &2.417&1.170  \\
        &720    & 0.436          & 0.424       &4.292&1.541  &1.128&0.718  &3.097&1.334 \\
 \hline 
  % \textbf{Avg} &   & 0.338          & 0.368    &1.716&0.903  &0.642&0.500   &2.010&1.034   \\
 %\hline
Weather &96         & 0.319          & 0.369     &0.351&0.410   &0.191&0.236     &0.367&0.407   \\
        &192        & 0.295          & 0.353        &0.712&0.603  &0.284&0.316   &0.459&0.478   \\
        &336         & 0.346          & 0.385      &0.422&0.441   &0.296&0.320    &0.671&0.597    \\
        &720        & 0.559          & 0.524       &1.026&0.737   &0.351&0.386    &0.645&0.604   \\
 \hline 
  % \textbf{Avg} &    & 0.379          & 0.407    &0.627&0.547  &0.280&0.314   &0.535&0.521   \\
 %\hline
Electricity  &96     & 0.216          & 0.335      &0.336&0.419  &0.170&0.271  &0.311&0.399  \\
               &192     & 0.274          & 0.363      &0.353&0.436  &0.182&0.284  &0.334&0.413  \\
               &336     & 0.243          & 0.353       &0.359&0.442  &0.199&0.299  &0.355&0.427  \\
               &720    & 0.287          & 0.372      &0.401&0.461  &0.226&0.324  &0.326&0.403  \\
 \hline 
  % \textbf{Avg} &   & 0.255          & 0.355   &0.362&0.439  &0.194&0.294   &0.331&0.410  \\
 %\hline
Traffic  &96      & 0.668          & 0.400       &0.741&0.423  &0.622&0.346   &0.723&0.405 \\
        &192      & 0.658          & 0.417        &0.766&0.434  &0.645&0.355  &0.710&0.390  \\
        &336      & 0.667          & 0.414     &0.893&0.506  &0.651&0.354  &0.697&0.384 \\
        &720      & 0.650          & 0.403        &1.049&0.587  &0.676&0.369  &0.707&0.387 \\
 \hline 
 %  \textbf{Avg} &   & 0.661          & 0.408   &0.862&0.487  &0.648&0.356   &0.709&0.391\\
% \hline
Exchange  &96       & 0.149& 0.279     &0.935&0.778  &0.141&0.262  &0.994&0.799  \\
        &192        & 0.285& 0.389     &1.116&0.843  &0.216&0.335  &1.456&0.978  \\
        &336        & 0.969& 0.728     &1.504&0.985  &0.493&0.510  &1.861&1.127  \\
        &720        & 1.112& 0.820     &2.932&1.414  &1.170&0.800  &1.835&1.149 \\
 \hline 
  % \textbf{Avg} &   & 0.628& 0.554     &1.621&1.005  &0.505&0.476   &1.536&1.013   \\
 %\hline
\end{tabular}
\label{tab1}
\end{table*}

\begin{table*}[hbt!]
\centering
\scriptsize
\captionsetup{justification=centering}
\caption{Measure of error coefficients on multivariate long-term forecasting results with different prediction lengths (96,192,336,720) (continue)}
\begin{tabular}{|p{2cm} |c |c c |c c |c c |c c | c c |}
\hline
Models & &\multicolumn{2}{c}{ATFNet}  &\multicolumn{2}{c}{MICN} &\multicolumn{2}{c}{Pyraformer} &\multicolumn{2}{c}{\textbf{EDformer}} \\
\hline
Database &Metric  &MSE&MAE &MSE&MAE &MSE&MAE   &MSE&MAE\\
\hline
ETTh1 &96       &0.418&0.442   &0.413 &0.442   &0.644 &0.597    &0.404 &0.415\\
    &192        &0.487&0.493   &0.451 &0.462   &0.843 &0.714    &0.492 &0.481\\
    &336        &0.516&0.522   &0.556 &0.528   &1.091 &0.841    &0.544 &0.519\\
    &720        &0.640&0.597   &0.658 &0.607   &1.005 &0.802    &0.628 &0.559\\
 \hline 
 %  \textbf{Avg} &   &0.515&0.513  &0.519 &0.509   &0.895 &0.738   &0.594&0.537\\
 %\hline
ETTh2 &96       &0.183&0.288   &0.303 &0.364   &1.380 &0.917  &0.386&0.417\\
    &192        &0.226&0.342   &0.403 &0.446   &5.927 &1.917  &0.521&0.490\\
    &336        &0.261&0.363   &0.603 &0.550   &4.515 &1.816  &0.621&0.562\\
    &720        &0.339&0.426   &1.106 &0.852   &4.367 &1.798  &0.625&0.554\\
 \hline 
 %  \textbf{Avg} &   &0.252&0.354  &0.603 &0.553   &4.047 &1.625  &0.538&0.505\\
 %\hline
ETTm1  &96      &0.376&0.402   &0.308 &0.360   &0.593 &0.520  &0.338&0.370\\
    &192        &0.427&0.432   &0.343 &0.384   &0.626 &0.547  &0.367&0.392\\
    &336        &0.480&0.464   &0.395 &0.411   &0.719 &0.616  &0.426&0.446\\
    &720        &0.551&0.513   &0.427 &0.434   &0.958 &0.732  &0.560&0.518\\
 \hline 
  % \textbf{Avg} &   &0.458&0.452  &0.368 &0.397   &0.724 &0.603   &0.422&0.431\\
 %\hline
ETTm2  &96      &0.118&0.231   &0.169 &0.268   &0.435 &0.507  &0.208&0.294\\
    &192        &0.147&0.258   &0.247 &0.333   &0.730 &0.673  &0.529&0.482\\
    &336        &0.182&0.288   &0.290 &0.351   &1.201 &0.845  &0.712&0.588\\
    &720        &0.231&0.327   &0.417 &0.434   &3.625 &1.415  &0.726&0.597\\
 \hline 
  % \textbf{Avg} &   &0.169&0.276  &0.281 &0.347   &1.497 &0.869   &0.543&0.479\\
 %\hline
Weather &96      &0.158&0.208   &0.178 &0.249   &0.203 &0.285  &0.219&0.261\\
    &192        &0.199&0.248   &0.243 &0.269   &0.219 &0.304  &0.252&0.294\\
    &336        &0.248&0.287   &0.278 &0.338   &0.294 &0.368  &0.325&0.336\\
    &720        &0.312&0.336   &0.320 &0.360   &0.388 &0.407  &0.423&0.402\\
 \hline 
   %\textbf{Avg} &   &0.229&0.270  &0.254 &0.304   &0.276 &0.341   &0.304&0.323\\
 %\hline
Electricity  &96      &0.136&0.230   &0.157 &0.266   &0.279 &0.374  &0.169&0.268\\
    &192              &0.151&0.248   &0.175 &0.287   &0.294 &0.390  &0.176&0.271\\
    &336              &0.169&0.265   &0.200 &0.308   &0.299 &0.395  &0.189&0.286\\
    &720              &0.207&0.299   &0.228 &0.338   &0.298 &0.386  &0.254&0.341\\
 \hline 
  % \textbf{Avg} &   &0.166&0.261  &0.190 &0.299   &0.292 &0.386   &0.197&0.292\\
% \hline
Traffic  &96    &0.380&0.264   &0.473 &0.293   &0.702 &0.404  &0.465&0.325\\
    &192        &0.399&0.275   &0.483 &0.298   &0.678 &0.387  &0.446&0.314\\
    &336        &0.419&0.280   &0.491 &0.303   &0.689 &0.390  &0.456&0.326\\
    &720        &0.441&0.299   &0.559 &0.327   &0.705 &0.410  &0.504&0.361\\
 \hline 
  % \textbf{Avg} &   &0.409&0.279  &0.502 &0.305   &0.693 &0.397   &0.467&0.332\\
 %\hline
Exchange  &96      &0.114&0.251      &0.278 &0.315   &0.533 &0.590  &0.112&0.247\\
    &192           &0.465&0.511      &0.392 &0.441   &0.960 &0.774  &0.470&0.527\\
    &336           &0.559&0.603      &0.551 &0.592   &1.191 &0.861  &0.691&0.651\\
    &720           &0.609&0.711      &0.612 &0.683   &1.670 &1.040  &0.875&0.912\\
 \hline 
  % \textbf{Avg} &   &0.436&0.519  &0.458 &0.507   &1.088 &0.816   &0.537&0.584\\
 %\hline
\end{tabular}
\label{tab2}
\end{table*}

\begin{table*}[hbt!]
\centering
\scriptsize
\captionsetup{justification=centering}
\caption{Comparison of average error coefficients on multivariate long-term forecasting result}
\begin{tabular}{|p{1.1cm}|c c |c c |c c |c c | c c | c c |c c |c c |}
\hline
Models    &\multicolumn{2}{c}{Autoformer} &\multicolumn{2}{c}{Informer} &\multicolumn{2}{c}{NS-Trans} &\multicolumn{2}{c}{Reformer} &\multicolumn{2}{c}{ATFNet}  &\multicolumn{2}{c}{MICN} &\multicolumn{2}{c}{PatchTST} &\multicolumn{2}{c}{\textbf{EDformer}} \\
\hline
Database  &MSE&MAE &MSE&MAE &MSE&MAE   &MSE&MAE &MSE&MAE &MSE&MAE &MSE&MAE   &MSE&MAE\\
 \hline 
ETTh1  & \textcolor{red}{0.504}& \textcolor{blue}{0.492}     &1.058&0.808  &0.609&0.541   &1.019&0.763   &0.515&0.513  &0.519 &0.509   &0.895 &0.738   &0.594&0.537\\
 \hline
ETTh2  &0.447& 0.463      &4.665&1.771  &0.567&0.509   &2.604&1.257  &\textcolor{red}{0.252}&\textcolor{blue}{0.354}  &0.603 &0.553   &4.047 &1.625  &0.538&0.505\\
 \hline
ETTm1   & 0.571& 0.513    &0.890&0.701  &0.521&0.472   &1.021&0.731   &0.458&0.452  &0.368 &0.397   &0.724 &0.603   &\textcolor{red}{0.422}&\textcolor{blue}{0.431}\\
 \hline
ETTm2  &\textcolor{red}{0.338}&\textcolor{blue}{0.368}    &1.716&0.903  &0.642&0.500   &2.010&1.034   &0.169&0.276  &0.281 &0.347   &1.497 &0.869   &0.543&0.479\\
 \hline
Weather  & 0.379 & 0.407    &0.627&0.547  &0.280&0.314   &0.535&0.521   &\textcolor{red}{0.229}&\textcolor{blue}{0.270}  &0.254 &0.304   &0.276 &0.341   &0.304&0.323\\
 \hline
Electricity   & 0.255& 0.355   &0.362&0.439  &0.199&0.294   &0.331&0.410   &0.201&0.261  &0.198&0.299   &0.292 &0.386   &\textcolor{red}{0.195}&\textcolor{blue}{0.292}\\
 \hline
Traffic   & 0.661& 0.408   &0.862&0.487  &0.648&0.356   &0.709&0.391    &0.479&0.339  &0.502 &0.345   &0.693 &0.397   &\textcolor{red}{0.467}&\textcolor{blue}{0.332}\\
 \hline
Exchange  & 0.628& 0.554     &1.621&1.005  &0.505&\textcolor{blue}{0.476}   &1.536&1.013    &\textcolor{red}{0.436}&0.519  &0.458&0.507   &1.088 &0.816   &0.537&0.584\\
 \hline
  \textbf{\# of Total Wins}  &2&2  &0 &0   &0 &1   &0&0   &\textcolor{red}{3}&2  &0 &0   &0 &0   &\textcolor{red}{3}&\textcolor{blue}{3}\\
 \hline
\end{tabular}
\label{avgtab}
\end{table*}

\begin{table*}[hbt!]
\centering
\scriptsize
\captionsetup{justification=centering}
\caption{Comparison of Standard deviations on multivariate long-term forecasting result with prediction horizons (96,192,336,720)}
\begin{tabular}{|c|cc|cc|cc|cc|cc|cc|cc|cc|}
\hline
Models      & \multicolumn{2}{c|}{Autoformer}    & \multicolumn{2}{c|}{Informer}      & \multicolumn{2}{c|}{NS-Trans}      & \multicolumn{2}{c|}{Reformer}      & \multicolumn{2}{c|}{ATFNet}        & \multicolumn{2}{c|}{MICN}          & \multicolumn{2}{c|}{Pyraformer}    & \multicolumn{2}{c|}{EDformer}      \\ \hline
Database    & \multicolumn{1}{c|}{MSE}   & MAE   & \multicolumn{1}{c|}{MSE}   & MAE   & \multicolumn{1}{c|}{MSE}   & MAE   & \multicolumn{1}{c|}{MSE}   & MAE   & \multicolumn{1}{c|}{MSE}   & MAE   & \multicolumn{1}{c|}{MSE}   & MAE   & \multicolumn{1}{c|}{MSE}   & MAE   & \multicolumn{1}{c|}{MSE}   & MAE   \\ \hline
ETTh1       & \multicolumn{1}{c|}{0.017} & 0.016 & \multicolumn{1}{c|}{0.212} & 0.123 & \multicolumn{1}{c|}{0.140} & 0.081 & \multicolumn{1}{c|}{0.166} & 0.068 & \multicolumn{1}{c|}{0.080} & 0.056 & \multicolumn{1}{c|}{0.095} & 0.064 & \multicolumn{1}{c|}{0.170} & 0.093 & \multicolumn{1}{c|}{0.081} & 0.053 \\ \hline
ETTh2       & \multicolumn{1}{c|}{0.042} & 0.033 & \multicolumn{1}{c|}{1.247} & 0.284 & \multicolumn{1}{c|}{0.058} & 0.034 & \multicolumn{1}{c|}{0.233} & 0.071 & \multicolumn{1}{c|}{0.057} & 0.049 & \multicolumn{1}{c|}{0.064} & 0.184 & \multicolumn{1}{c|}{1.655} & 0.403 & \multicolumn{1}{c|}{0.097} & 0.058 \\ \hline
ETTm1       & \multicolumn{1}{c|}{0.030} & 0.014 & \multicolumn{1}{c|}{0.239} & 0.126 & \multicolumn{1}{c|}{0.075} & 0.043 & \multicolumn{1}{c|}{0.159} & 0.065 & \multicolumn{1}{c|}{0.065} & 0.041 & \multicolumn{1}{c|}{0.045} & 0.027 & \multicolumn{1}{c|}{0.142} & 0.081 & \multicolumn{1}{c|}{0.085} & 0.057 \\ \hline
ETTm2       & \multicolumn{1}{c|}{0.058} & 0.033 & \multicolumn{1}{c|}{1.529} & 0.400 & \multicolumn{1}{c|}{0.312} & 0.143 & \multicolumn{1}{c|}{0.874} & 0.261 & \multicolumn{1}{c|}{0.042} & 0.035 & \multicolumn{1}{c|}{0.089} & 0.059 & \multicolumn{1}{c|}{1.257} & 0.342 & \multicolumn{1}{c|}{0.208} & 0.125 \\ \hline
Weather     & \multicolumn{1}{c|}{0.125} & 0.068 & \multicolumn{1}{c|}{0.266} & 0.131 & \multicolumn{1}{c|}{0.057} & 0.053 & \multicolumn{1}{c|}{0.127} & 0.082 & \multicolumn{1}{c|}{0.057} & 0.047 & \multicolumn{1}{c|}{0.052} & 0.046 & \multicolumn{1}{c|}{0.073} & 0.048 & \multicolumn{1}{c|}{0.078} & 0.052 \\ \hline
Electricity & \multicolumn{1}{c|}{0.027} & 0.013 & \multicolumn{1}{c|}{0.023} & 0.015 & \multicolumn{1}{c|}{0.021} & 0.019 & \multicolumn{1}{c|}{0.015} & 0.011 & \multicolumn{1}{c|}{0.026} & 0.025 & \multicolumn{1}{c|}{0.026} & 0.026 & \multicolumn{1}{c|}{0.008} & 0.007 & \multicolumn{1}{c|}{0.033} & 0.029 \\ \hline
Traffic     & \multicolumn{1}{c|}{0.007} & 0.007 & \multicolumn{1}{c|}{0.121} & 0.065 & \multicolumn{1}{c|}{0.019} & 0.008 & \multicolumn{1}{c|}{0.009} & 0.008 & \multicolumn{1}{c|}{0.022} & 0.012 & \multicolumn{1}{c|}{0.033} & 0.013 & \multicolumn{1}{c|}{0.011} & 0.009 & \multicolumn{1}{c|}{0.021} & 0.017 \\ \hline
Exchange    & \multicolumn{1}{c|}{0.417} & 0.225 & \multicolumn{1}{c|}{0.783} & 0.247 & \multicolumn{1}{c|}{0.405} & 0.207 & \multicolumn{1}{c|}{0.352} & 0140  & \multicolumn{1}{c|}{0.193} & 0.170 & \multicolumn{1}{c|}{0.131} & 0.141 & \multicolumn{1}{c|}{0.410} & 0.162 & \multicolumn{1}{c|}{0.284} & 0.239 \\ \hline
\end{tabular}
\label{stdlong}
\end{table*}

\begin{table*}[hbt!]
\centering
\scriptsize
\captionsetup{justification=centering}
\caption{Comparison of average error coefficients on multivariate short-term forecasting result with prediction horizon 24 and fixed look-back 96}
\begin{tabular}{|p{1.1cm}|c c |c c |c c |c c | c c | c c |c c |c c |}
\hline
Models    &\multicolumn{2}{c}{Autoformer} &\multicolumn{2}{c}{Informer} &\multicolumn{2}{c}{NS-Trans} &\multicolumn{2}{c}{Reformer} &\multicolumn{2}{c}{ATFNet}  &\multicolumn{2}{c}{MICN} &\multicolumn{2}{c}{PatchTST} &\multicolumn{2}{c}{\textbf{EDformer}} \\
\hline
Database                  & MSE                         & MAE                         & MSE                               & MAE                               & MSE                        & MAE                        & MSE                                & MAE                                & MSE                       & MAE                       & MSE                      & MAE                      & MSE                                 & MAE                                & MSE     & \multicolumn{1}{c|}{MAE}    \\ \hline
PEMS03                    & 0.469                       & 0.508                       & 0.111                             & 0.226                             & \textcolor{red}{0.106}                      & \textcolor{blue}{0.216}                      & 0.115                              & 0.231                              & 0.146                     & 0.262                     & 0.141                    & 0.275                    & 0.179                               & 0.279                              & 0.187   & \multicolumn{1}{c|}{0.307}  \\ \hline
PEMS04                    & 0.346                       & 0.437                       & \textcolor{red}{0.092}                            & \textcolor{blue}{0.198}                            & 0.103                      & 0.215                      & 0.100                              & 0.210                              & 0.158                     & 0.278                     & 0.195                    & 0.327                    &0.221                               & 0.316                              & 0.191   & \multicolumn{1}{c|}{0.311}  \\ \hline
PEMS07                    & 0.223                       & 0.351                      & 0.108                             & 0.216                             & \textcolor{red}{0.100}                      & \textcolor{blue}{0.211}                      & 0.108                              & 0.219                              & 0.144                     & 0.264                     & 0.123                    & 0.253                    & 0.189                               & 0.294                             & 0.165   & \multicolumn{1}{c|}{0.291}  \\ \hline
PEMS08                    & 0.485                       & 0.537                       & 0.128                             & 0.249                             & \textcolor{red}{0.116}                      & \textcolor{blue}{0.229}                      & 0.137                              & 0.259                              & 0.141                     & 0.261                     & 0.177                    & 0.312                    & 0.218                               & 0.295                              & 0.214   & \multicolumn{1}{c|}{0.336}  \\ \hline

%\textbf{\# of Total Wins}  &0&0  &0 &0   &0 &0   &0&0   &\textcolor{red}{0}&0  &0 &0   &0 &0   &\textcolor{red}{0}&\textcolor{blue}{0}\\
 \hline
\end{tabular}
\label{avgtabshort}
\end{table*}

\begin{table*}[hbt!]
\centering
\scriptsize
\captionsetup{justification=centering}
\caption{Short-term forecasting results in the M4 dataset with a single variate. All prediction lengths are in [6, 48].  The values highlighted in red colour indicate the best results for each row.}
\adjustbox{max width=\textwidth}{
\begin{tabular}{|l|c|c|c|c|c|c|c|}
\textbf{Metric}                 & \textbf{Category} & \textbf{EDformer}                                                  & \textbf{iTransformer} & \textbf{Reformer}                                                   & \textbf{NS-Trans}         & \textbf{Informer}         & \textbf{Autoformer}\\ \hline
\multirow{5}{*}{\textbf{sMAPE}} & Yearly            & \textcolor{red}{14.259}    & 14.409  &14.548  &15.833   & 15.215  &16.909  \\
                                & Quarterly         & 11.407  & \textcolor{red}{10.777}  &11.922   &12.366    & 12.696  &14.445  \\
                                & Monthly           & 15.558  & 16.650 &14.649   &\textcolor{red}{14.607}   & 15.210 &18.280  \\
                                & Others            & \textcolor{red}{5.222}   & 5.543   &6.694     &7.005   & 7.183  &6.676 \\ \hline
                                & \textbf{Average}  & \textcolor{red}{\textbf{13.796}} & \textbf{14.170}  & \textbf{14.192}                              & \textbf{14.201}   & \textbf{14.206}  & \textbf{16.464}\\ \hline
\multirow{5}{*}{\textbf{MAPE}}  & Yearly            &\textcolor{red}{17.558}    & 19.191 & 17.789     &20.485     & 19.837    &23.266  \\
                                & Quarterly         &13.006   & 12.871  &\textcolor{red}{12.737}      &14.490     & 14.969    &16.882   \\
                                & Monthly           & 18.318  & 20.144 &\textcolor{red}{15.830}       &16.988     &  17.972   &22.442   \\
                                & Others            &\textcolor{red}{7.142} & 7.750 &10.456          &10.459     & 10.469    &11.146    \\ \hline
                                & \textbf{Average}  & \textbf{16.409}                                                      & \textbf{17.560}       &\textcolor{red}{\textbf{14.971}}        & \textbf{16.689}   & \textbf{17.305}  & \textbf{20.732}\\ \hline 
\multirow{5}{*}{\textbf{MASE}}  & Yearly            &\textcolor{red}{3.158}    & 3.218     &3.232   &3.532     & 3.398   &3.761  \\
                                & Quarterly         &1.426    &\textcolor{red}{1.284}   &1.313      &1.519     & 1.561     &1.854  \\
                                & Monthly           &\textcolor{red}{1.189}   & 1.392  &1.262        & 1.177       &  1.217     &1.572  \\
                                & Others            &4.568    &\textcolor{red}{3.998}   &4.424       &4.691        &  4.937     &4.833  \\ \hline
                                & \textbf{Average}  & \textcolor{red}{\textbf{1.868}}   & \textbf{1.916} & \textbf{1.894}  & \textbf{1.910}      &  \textbf{1.987}    & \textbf{2.306}\\ \hline
\multirow{5}{*}{\textbf{OWA}}   & Yearly            &0.834   & 0.846  &\textcolor{red}{0.796}    & 0.929   & 0.893   &0.991  \\
                                & Quarterly         & 1.038  & \textcolor{red}{0.957}   & 0.975     &1.115    &  1.145   &1.332  \\
                                & Monthly           & 1.098  & 1.232   &\textcolor{red}{0.972}     &1.060   &  1.099    &1.373  \\
                                & Others            &1.375   & \textcolor{red}{1.214}  & 1.402     &1.502     &  1.534    &1.465   \\ \hline
                                & \textbf{Average}  & \textbf{0.997}  & \textbf{1.023}  & \textcolor{red}{\textbf{0.921}}  & \textbf{1.039}      &  \textbf{1.043}     & \textbf{1.210}\\ \hline
\end{tabular}
}
\label{table-resultsM4}
\end{table*}

\begin{table*}[hbt!]
\centering
\scriptsize
\caption{The cost time (sec) and speed (running time, s/iter) of some benchmark models on ETTh1 dataset}
\begin{tabular}{|c|ll|lc|cl|cc|}
\hline
Series Length & \multicolumn{2}{c}{96}                                                    & \multicolumn{2}{c}{192}                                                                    & \multicolumn{2}{c}{336}                                     & \multicolumn{2}{c}{720}  
\\
\hline
Models                            & Speed                               & Cost time                           & Speed                               &Cost time & \multicolumn{1}{l}{Speed}     & Cost time                   & \multicolumn{1}{l}{Speed} &Cost time \\
\hline
\textbf{EDformer}                       &\textcolor{red}{0.0050}                                &\textcolor{blue}{1.5344}                                    &\textcolor{red}{0.0053}                                    &\textcolor{blue}{1.5714}                              &\textcolor{red}{0.0055}                              &\textcolor{blue}{1.5811}                            &\textcolor{red}{0.0058}                          &\textcolor{blue}{1.5299}                              \\
\hline
Pyraformer                       & 0.0194                              & 5.3420                              & 0.0199                              &5.3802                        & 0.0199                        & 5.2862                      & 0.0202                    & 5.1609                        \\
\hline
Autoformer                        & 0.0402                              & 10.881                              & 0.0525                              & { 13.952}                        & 0.0719                        & 18.689                      & 0.1278                    & 28.331                        \\
\hline
Reformer                      & 0.0361                              & 9.7632                              & 0.0531                              &14.082                        & 0.0764                        & 19.823                      & 0.1451                    & 35.682                        \\
\hline
Informer                         & 0.0222                              & 6.0839                              & 0.0269                              &7.2604                        & 0.0349                        &9.1842                      & 0.0542                    &13.477                       \\
\hline
FEDformer                         & 0.1428                              & 38.693                              & 0.1524                              &40.891                        & 0.1614       & 42.312        & 0.2048                    & 51.016                        \\
\hline
\end{tabular}
\label{tabspeed}
\end{table*}

\begin{table*}[hbt!]
\centering
\scriptsize
\caption{Comparison of multivariate long-term and short-term forecasting results in terms of average execution time (sec)}
\begin{tabular}{|c|l|c|c|c|c|c|c|}
\hline
\multicolumn{1}{l}{}           & Datasets & \textbf{EDformer} & Autoformer & Informer & Reformer &Pyraformer &NS-Trans\\
\hline
\multirow{8}{*}{\begin{tabular}[c]{@{}c@{}}Execution Time (Sec) \\ (Total)\end{tabular}} 
                                     & ETTh1   & \textcolor{red}{0.831}      & 2.715     & 1.591    &1.928    & 1.541   &1.220\\
                                      & ETTh2   & \textcolor{red}{0.746}      & 4.242    & 1.551    & 1.936   & 1.544   &1.315\\
                                      & Weather   & \textcolor{red}{3.168}      & 12.462    &5.423    &6.711   & 5.245   &5.831\\
                                      & Exchange   & \textcolor{red}{0.514}      & 1.826    &0.928   & 1.114   & 0.884   &0.991\\
                                      \hline
& PEMS03   & \textcolor{red}{0.878}      & 1.162     & 1.150       & 1.190   & 1.177   &1.171\\
& PEMS04   & \textcolor{red}{0.848}      & 1.116    & 1.253     & 1.210  & 1.213    &1.163\\
& PEMS07   & \textcolor{red}{1.180}      & 1.302    & 1.287       & 1.390   & 1.316   &1.378\\
& PEMS08   & \textcolor{red}{0.703}      & 1.059    & 1.099      & 1.126   & 1.066   &1.078\\
\hline
\end{tabular}
\label{table-executionpems}
\end{table*}

\section{ABLATION STUDIES AND PERFORMANCE COMPARISONS}
The ablation study presented in Table \ref{table-ablation} shows the average performance of different configurations on seven datasets: Etth1, Ettm1, Weather, Electricity, and Traffic, over four prediction lengths (96, 192, 336, 720). We aim to show the effects of incorporating reverse-embedding operations and decomposition mechanisms into the forecasting model. The table compares the mean squared error (MSE) and mean absolute error (MAE) for three key configurations: using without decomposition, with decomposition, and reverse embedding in combination with decomposition. The results demonstrate that when average decomposition is employed (second and third rows), the model consistently outperforms the model without decomposition (first row) across all datasets. This highlights the significance of allowing the model to learn how to decompose the input time series rather than using a predefined approach. Moreover, the addition of reverse embedding further improves the results (third row), suggesting that reverse embedding, which reverses the whole seasonal signal embedding, provides crucial temporal information that enhances forecasting accuracy. Overall, the combination of average decomposition and reverse embedding leads to the most significant performance gains, making these features essential components in improving the model’s forecasting ability.

\begin{table}[hbt!]
\centering
\scriptsize
\caption{Ablation Study: Comparison of Multivariate Long-Term Forecasting for the Average Results Across Prediction Horizons {96, 192, 336, 720}}
\begin{tabular}{|c|c|c|c|c|c|c|}
\hline
\textbf{} & \textbf{\begin{tabular}[c]{@{}c@{}}Without \\ decompose\end{tabular}} & \textbf{\begin{tabular}[c]{@{}c@{}}With \\ decompose\end{tabular}} & \textbf{\begin{tabular}[c]{@{}c@{}}Reverse\\ Embedding\end{tabular}} & \textbf{MSE}     & \textbf{MAE}     \\ \hline
   & \checkmark & $-$ & $-$ & 1.010 & 0.810 \\
\textbf{Etth1}    & $-$ & \checkmark & $-$ & 0.789 & 0.697 \\
    & $-$ & \checkmark & \checkmark & \textcolor{red}{0.594} & \textcolor{blue}{0.537} \\ \hline
  %  & \checkmark & $-$ & $-$ & 3.027 & 1.484 \\
%\textbf{Etth2}    & $-$ & \checkmark & $-$ & 2.177 & 0.975 \\
 %   & $-$ & \checkmark & \checkmark & \textcolor{red}{0.538} & \textcolor{blue}{0.505} \\ \hline
   & \checkmark & $-$ & $-$ & 0.905 & 0.799 \\
\textbf{Ettm1}    & $-$ & \checkmark & $-$ & 0.769 & 0.638 \\
   & $-$ & \checkmark & \checkmark & \textcolor{red}{0.422} & \textcolor{blue}{0.431} \\ \hline
 %  & \checkmark & $-$ & $-$ & 1.  & 1. \\
%\textbf{Ettm2}    & $-$ & \checkmark & $-$ & 0.280  & 0.323 \\
%    & $-$ & \checkmark & \checkmark & \textcolor{red}{0.543}  & \textcolor{blue}{0.479} \\ \hline
    & \checkmark & $-$ & $-$ & 0.691  & 0.613 \\
\textbf{Weather}    & $-$ & \checkmark & $-$ & 0.601  & 0.552 \\
    & $-$ & \checkmark & \checkmark & \textcolor{red}{0.304}  & \textcolor{blue}{0.323} \\ \hline
    & \checkmark & $-$ & $-$ & 0.441  & 0.502 \\
\textbf{Electricity}    & $-$ & \checkmark & $-$ & 0.361  & 0.399 \\
    & $-$ & \checkmark & \checkmark & \textcolor{red}{0.195}  & \textcolor{blue}{0.292} \\ \hline
    & \checkmark & $-$ & $-$ & 0.697  & 0.381 \\
\textbf{Traffic}    & $-$ & \checkmark & $-$ & 0.601  & 0.349 \\
    & $-$ & \checkmark & \checkmark & \textcolor{red}{0.467}  & \textcolor{blue}{0.332} \\ \hline
\end{tabular}
\label{table-ablation}
\end{table}

\section{Explainability Analysis}
\label{explain}
In multivariate time series forecasting, model explainability is essential because it enables stakeholders to comprehend, believe, and analyse forecasts produced by intricate models. Understanding how each feature affects the prediction outcome is essential for diagnosing model behaviour, enhancing model robustness, and learning about variable dependencies because multivariate time series data is high-dimensional and involves multiple interdependent variables that affect the forecast \citep{yaprakdal2023multivariate}. Each prediction can be attributed to a different characteristic using some well-known feature explainability techniques like feature ablation, feature occlusion, integrated gradients, gradient-SHAP, and windowed feature importance in time. These techniques are briefly described below.
\begin{itemize}
    \item \textbf{Feature Ablation (FA)} is an explainability technique that systematically removes one feature at a time from the input data to assess its impact on the model's performance. By measuring the change in prediction accuracy or output, this method identifies the importance of individual features. It helps highlight which features contribute most significantly to the model's decisions, making it useful for understanding the behavior of complex models \citep{ismail2020benchmarking}.
    \item \textbf{Feature Occlusion (FO)} involves masking or replacing parts of the input data (e.g., setting a feature to zero) to observe the change in the model's output. By selectively occluding features or segments of the input, this method can identify which components are most influential in the prediction process. It is commonly used in time series and image data analysis to understand spatial or temporal feature importance \citep{ismail2020benchmarking}.
    \item \textbf{Integrated Gradients (IG)} is an attribution method that explains model predictions by accumulating gradients along a path from a baseline input (e.g., all zeros) to the actual input. It calculates the contribution of each feature by integrating the gradients of the model's output with respect to the input features. This technique ensures that the attributions are consistent and satisfy properties like completeness, making it a reliable approach for model interpretability \citep{wang2024gradient}.
    \item \textbf{Shapley Additive explanations (SHAP)} uses cooperative game theory to allocate feature priority, quantifying each variable's contribution across all conceivable feature subsets \citep{tripathy2022explaining}. In the context of a Transformer model for multivariate time series forecasting, SHAP can provide insights into \citep{saluja2021towards}:
\begin{itemize}
    \item Which time steps (lags) are most influential?
    \item Which variables (features) contribute most to the prediction?
    \item How do different values impact the forecast?
\end{itemize}
\textbf{Gradient SHAP (GS)} is a hybrid explainability technique that combines SHAP (SHapley Additive exPlanations) values with gradient-based methods. It uses stochastic sampling and integrates gradients over a range of input samples to approximate the Shapley values for each feature. This method provides robust feature attributions and is particularly effective for complex, non-linear models, offering insights into how each feature influences the predictions \citep{wang2024gradient}.
\item \textbf{Windowed Feature Importance in Time (WinIT)} is a feature removal based explainability approach. WinIT explicitly incorporates temporal dependencies by considering the relationship between consecutive observations of the same feature when calculating its importance score. Additionally, it accounts for the dynamic nature of feature importance over time by summarizing its significance across a window of previous time steps, thereby capturing temporal variations in the feature's influence \citep{leung2021temporal,sa_timeseries}.
\end{itemize}
Two metrics \citep{sa_timeseries} have been used here to measure the importance of these above explainability methods.\\
\textbf{A. Comprehensiveness:} It evaluates how important a subset of features is for the model’s prediction. It measures the impact of removing these key features on the model’s output. The idea is that if the identified important features are truly contributing to the prediction, then omitting them should lead to a significant drop in the model’s confidence or prediction score. A higher comprehensiveness score indicates better feature attribution, as it confirms that the model relies on these features to make its decisions. It helps in validating the quality of the explainability technique by checking if the most influential features indeed hold crucial information for the model. Table \ref{comp1} presents a comparison of 'Comprehensiveness' scores (average MAE) for some state-of-the-art explainability methods on the Electricity dataset. Table \ref{comp2} presents a comparison of 'Comprehensiveness' scores (average MSE) for some state-of-the-art explainability methods on the Electricity dataset. It is evident that our proposed EDformer model registers the highest number of wins in terms of 'Comprehensiveness' compared to other methods.\\
\textbf{B. Sufficiency:} It assesses whether the identified features alone are sufficient to maintain the prediction outcome. It does this by measuring the performance of the model when only the key attributed features are retained, while the rest are removed or masked. If the model's predictions remain consistent, it indicates that these features are adequate for making accurate predictions. For the sufficiency metric, a higher value is considered better. The sufficiency score helps determine if the feature attribution is complete, ensuring that the identified features not only contribute significantly but also capture enough information to make reliable predictions. Table \ref{suff1} presents a comparison of 'Sufficieny' scores (average MAE) for some state-of-the-art explainability methods on the Electricity dataset. Table \ref{suff2} presents a comparison of 'Sufficieny' scores (average MSE) for some state-of-the-art explainability methods on the Electricity dataset. It is evident that our proposed EDformer model registers the highest number of wins in terms of 'Sufficiency' compared to other methods.

\begin{table*}[hbt!]
\centering
\scriptsize
\caption{Comparison of Comprehensiveness (avg. of MAE) among explainability methods on Electricity dataset for prediction length 24. The highest value is marked by red colour.}
\begin{tabular}{|c|c|c|c|c|c|c|}
\hline
Method & Autoformer & Informer & NS-Trans & Reformer & MICN  & EDformer  \\ \hline
FA     & \textcolor{red}{12.1}           & 4.1         & 9.8         & 3.4         & 12.0             & 11.6         \\ \hline
FO     &  14.1          & 6.1         & 12.6         & 4.3         & 11.0            & \textcolor{red}{14.7}       \\ \hline
IG     &  13.0          & 12.1         & \textcolor{red}{14.9}         & 8.5         & 13.6           & 14.4        \\ \hline
GS     & \textcolor{red}{11.0}          &  5.3        &  7.4        & 5.8         & 10.4             & 8.2        \\ \hline
WinIT   &  14.1         &  6.5        & 12.4         &  4.8        & 12.6           & \textcolor{red}{14.3}        \\ \hline
\textbf{\# of Total Wins}  &2   &0   & 1  &0   &0   &2\\
\hline
\end{tabular}
\label{comp1}
\end{table*}

\begin{table*}[hbt!]
\centering
\scriptsize
\caption{Comparison of Comprehensiveness (avg. of MSE) among explainability methods on Electricity dataset for prediction length 24. The highest value is marked by red colour.}
\begin{tabular}{|c|c|c|c|c|c|c|}
\hline
Method & Autoformer & Informer & NS-Trans & Reformer & MICN  & EDformer \\ \hline
FA     & 9.68           & 1.47         & 8.4         & 1.0         & 10.5             & \textcolor{red}{10.7}         \\ \hline
FO     &  12.8          & 3.5         & 12.2         & 1.9         & 11.2             & \textcolor{red}{15.8}       \\ \hline
IG     &  11.1          & 12.0         & \textcolor{red}{16.9}         & 6.3         & 12.9             & 15.2         \\ \hline
GS     &   8.2        &  2.7        & 5.1         &  3.3        & \textcolor{red}{8.5}             & 5.9        \\ \hline
WinIT   &   12.7        &  4.0        & 11.7         &  2.2        & 11.3            & \textcolor{red}{14.8}        \\ \hline
\textbf{\# of Total Wins}  &0   &0   &1   &0   &1   &3\\
\hline
\end{tabular}
\label{comp2}
\end{table*}

\begin{table*}[hbt!]
\centering
\scriptsize
\caption{Comparison of Sufficiency (avg. of MAE) among explainability methods on Electricity dataset for prediction length 24. The highest value is marked by red colour.}
\begin{tabular}{|c|c|c|c|c|c|c|}
\hline
Method & Autoformer & Informer & NS-Trans & Reformer & MICN  & EDformer \\ \hline
FA     & 14.5           & 15.3         & 15.5         &  12.2        & 11.1            &  \textcolor{red}{17.2}       \\ \hline
FO     & 14.1           & 14.9         & 13.5         &  11.7        & 9.8              & \textcolor{red}{15.0}       \\ \hline
IG     & 16.3         &  18.9        &  21.2        &  15.5        &  16.3             & \textcolor{red}{28.6}         \\ \hline
GS     & 15.3           & 14.8         & 17.7         & 12.1         &  14.6       & \textcolor{red}{21.9}        \\ \hline
WinIT   & 14.1          &  14.6        & 13.6         & 11.5         & 9.8            & \textcolor{red}{15.8}        \\ \hline
\textbf{\# of Total Wins}  &0   &0   &0   &0   &0   &5\\
\hline
\end{tabular}
\label{suff1}
\end{table*}

\begin{table*}[hbt!]
\centering
\scriptsize
\caption{Comparison of Sufficiency (avg. of MSE) among explainability methods on Electricity dataset for prediction length 24. The highest value is marked by red colour.}
\begin{tabular}{|c|c|c|c|c|c|c|}
\hline
Method & Autoformer & Informer & NS-Trans & Reformer & MICN  & EDformer \\ \hline
FA     &  13.3          & 16.2         & 16.2         &  10.5        & 8.7       & \textcolor{red}{19.8}        \\ \hline
FO     &  12.6          & 15.4         & 12.8         &  9.7        & 7.0        & \textcolor{red}{15.5}       \\ \hline
IG     &  16.4          &  22.8        & 27.3         & 16.3         & 17.1       & \textcolor{red}{51.7}        \\ \hline
GS     &  14.6         & 14.7         & 20.1         &  10.2        & 14.0        & \textcolor{red}{31.2}       \\ \hline
WinIT   &  12.6          & 15.0         & 13.1         & 9.4         & 7.1         & \textcolor{red}{16.9}      \\ \hline
\textbf{\# of Total Wins}  &0   &0   &0  & 0  &0   &5\\
\hline
\end{tabular}
\label{suff2}
\end{table*}

\begin{figure}[hbt!]
\centering
\includegraphics[scale=0.3]{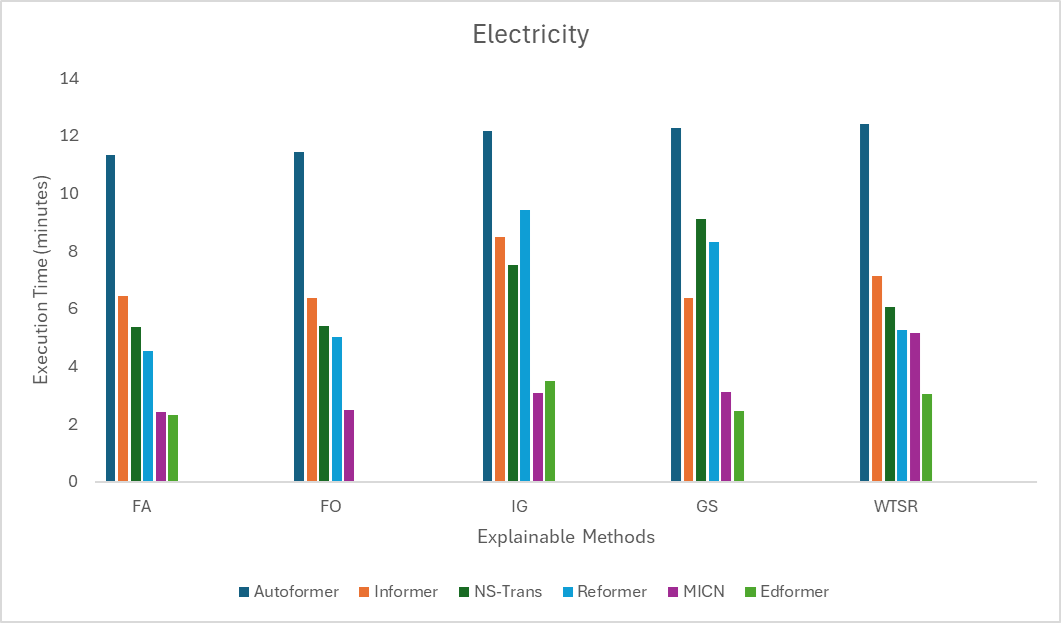}
\caption{Comparison on execution time (minutes) for model interpretability on Electricity dataset}
\label{exec_inter}
\end{figure}
Figure \ref{exec_inter} shows the total execution time (minutes) taken for interpretability assessment of each model on the electricity dataset. 

\subsection{Discussion on Explainability and Feature Importance}
Table \ref{comp1} to Table \ref{suff1} evaluate the comprehensiveness and sufficiency of explainability methods across six forecasting models (Autoformer, Informer, NS-Trans, Reformer, MICN, and EDformer) for the Electricity (ECL) dataset with a prediction length of 24. Comprehensiveness measures the reduction in predictive accuracy when the most important features identified by an explainability method are removed, while sufficiency assesses the model's performance when only these key features are retained. The highest scores (marked in red) indicate the explainability method’s ability to capture the most impactful features effectively. The analysis reveals that EDformer consistently achieves the highest scores in both metrics, particularly excelling in Sufficiency. This suggests that EDformer benefits the most from the features identified by the explainability methods, reinforcing its robustness in utilizing relevant information. Conversely, models like Informer and Reformer have fewer "wins," indicating potential room for improvement in aligning their performance with explainability insights. The use of this explainability analysis ensures that models are not treated as "black boxes." It helps identify the significance of input features, improves transparency, and fosters trust in AI predictions. This is particularly crucial for datasets like 'Electricity', where decisions can impact critical energy management systems. By understanding feature importance, stakeholders can optimize model inputs, enhance interpretability, and ensure alignment with real-world domain knowledge.
\par
\begin{figure}[hbt!]
    \centering
    \vspace{1em}
    \begin{subfigure}{0.4\textwidth}
        \includegraphics[width=\textwidth]{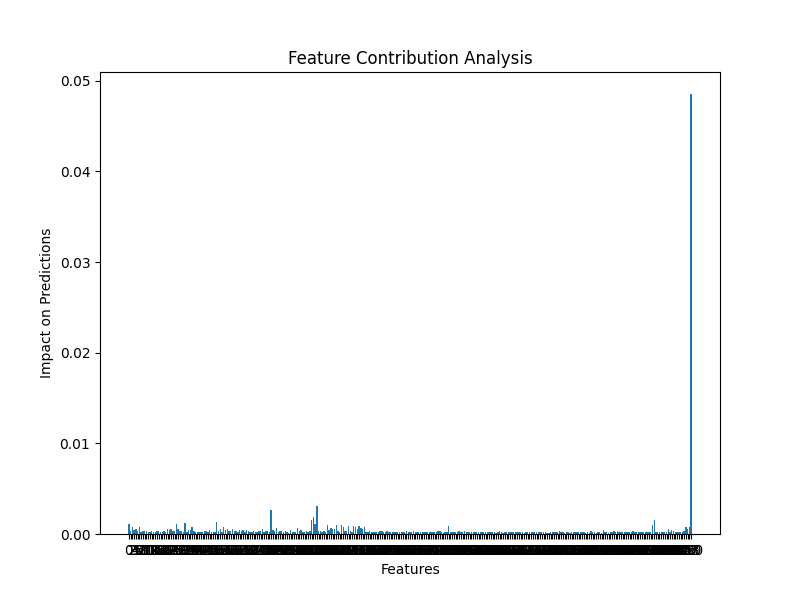}
        \caption{EDformer (GS)}
    \end{subfigure}
    \hfill
    \begin{subfigure}{0.4\textwidth}
        \includegraphics[width=\textwidth]{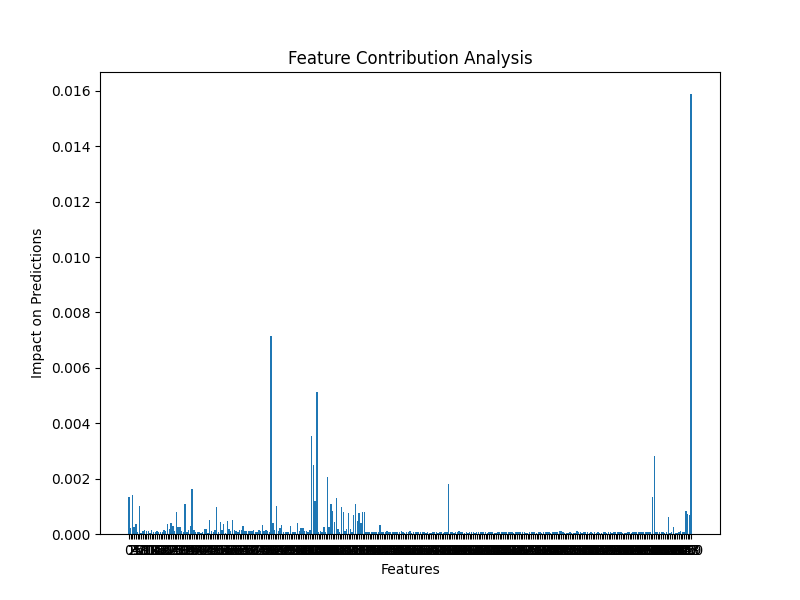}
        \caption{EDformer (FA)}
    \end{subfigure}
    
    \caption{Feature importance of EDformer model on the electricity (ECL) dataset (prediction length: 24) using Gradient Shap (GS), and Feature ablation (FA) methods.}
    \label{feature-importance}
\end{figure}

\begin{figure}[hbt!]
    \centering
    \vspace{1em}
    \begin{subfigure}{0.4\textwidth}
        \includegraphics[width=\textwidth]{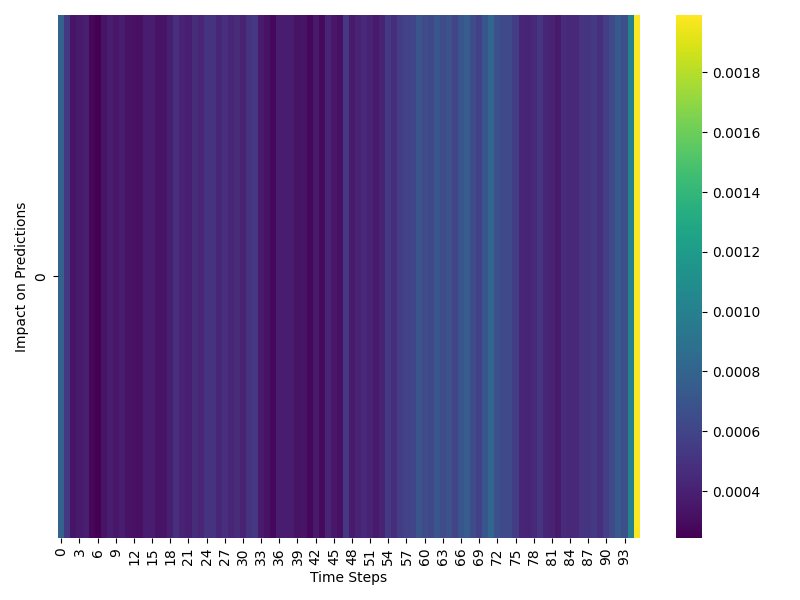}
        \caption{EDformer (GS)}
    \end{subfigure}
    \hfill
    \begin{subfigure}{0.4\textwidth}
        \includegraphics[width=\textwidth]{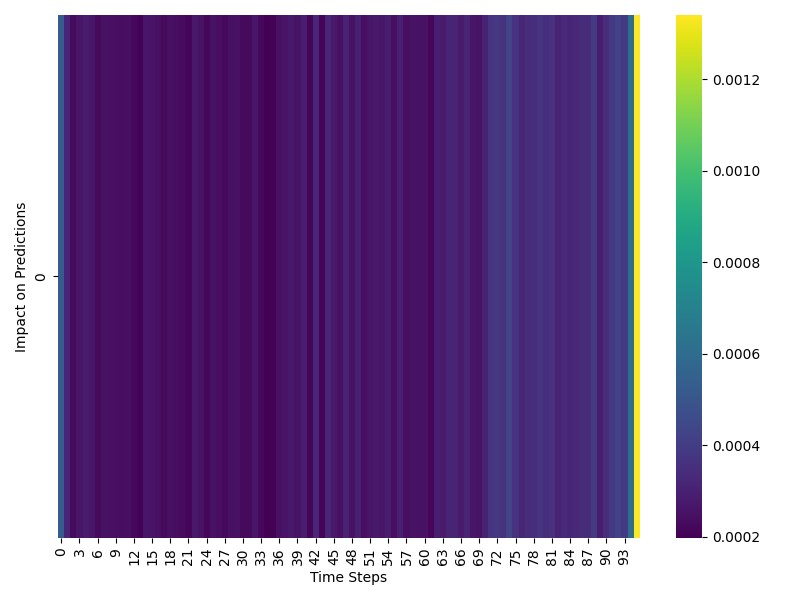}
        \caption{EDformer (FA)}
    \end{subfigure}
    
    \caption{Feature saliency of EDformer model on the electricity (ECL) dataset (prediction length: 24) using Gradient Shap (GS), and Feature ablation (FA) methods.}
    \label{feature-saliency}
\end{figure}

 The feature importance analysis in multivariate time series forecasting models provides valuable insights into the contribution of different input variables toward predicting the target outcomes. Using methods such as Gradient SHAP (GS), and feature ablation (FA), the significance of electricity (ECL) dataset features is evaluated for the EDformer model. The visual representations of Fig. \ref{feature-importance} highlight variations in feature importance across methods and the EDformer model. Figure \ref{feature-importance} highlights that the electricity consumption/load for future time steps (OT) is the most influential feature in the overall prediction of the EDformer model, as it directly represents the target class. Additionally, other features such as seasonality, temperature, time of day, and humidity also exhibit notable impacts on the prediction results. The saliency analysis depicted in Fig.\ref{feature-saliency}, illustrates the predictive relevance of individual features in the multivariate time-series models applied to the electricity (ECL) dataset. The saliency maps show variation in feature importance across models, indicating diverse sensitivity patterns. This analysis underscores the potential of interpretability methods in guiding model selection and enhancing understanding of process dynamics in multivariate time-series forecasting tasks.

\section{Conclusions}
\label{conclusions}
In this paper, EDformer, an architecture for multivariate time series forecasting that integrates decomposition into seasonal and trend components with a reverse embedding process, followed by an encoding-based forecaster, has been introduced. EDformer consistently achieves state-of-the-art performance across a wide range of benchmarks, showcasing its generality and robustness for long-term and short-term forecasting tasks. Moreover, EDformer has demonstrated state-of-the-art runtime efficiency, reduced cost time, and significant speed-up compared to other models. This paper delves into advanced model explainability techniques—such as feature ablation, feature occlusion, integrated gradients, gradient-SHAP, and windowed feature importance over time—within the realm of time series forecasting. It aims to identify which black-box algorithms derive the greatest benefits from these explainability methods, thereby highlighting their robustness in effectively utilizing critical features and improving model interpretability. Detailed visualizations and ablations are included to provide insights into our architecture. In the future, this model will be tested for some real-life time series datasets.

\section*{Acknowledgements}
This work was partially supported by the 'Resurssmarta Processor (RSP)', the Wallenberg AI, Autonomous Systems and Software Program (WASP), and the Wallenberg Initiative Materials Science for Sustainability (WISE), all funded by the Knut and Alice Wallenberg Foundation. 

%\clearpage

\bibliographystyle{unsrt}
\bibliography{reference}

\vfill

\end{document}